\documentclass[letterpaper]{article} 
\usepackage[preprint]{aaai2027}
\usepackage[hyphens]{url}  
\usepackage{graphicx} 
\usepackage{natbib}  
\usepackage{caption} 
\usepackage{algorithm}
\usepackage{algorithmic}
\graphicspath{{Figures/}}
\usepackage{placeins}
\usepackage{amsmath,amssymb}
\usepackage{newfloat}
\usepackage{listings}
\usepackage{multirow}
\usepackage{array}
\usepackage{tabularx}
\floatstyle{ruled}
\newfloat{listing}{tb}{lst}{}
\floatname{listing}{Listing}

\usepackage{booktabs}

\title{Beyond Representational Similarity: Source-Conditioned Description-Length Gain for Generative Plagiarism Detection and Candidate Source Reranking}
\author{
    Peijia Guo\textsuperscript{\rm 1}\equalcontrib,
    Wenxuan Xie\textsuperscript{\rm 1,2}\equalcontrib,
    ZiGuang Li\textsuperscript{\rm 1}\equalcontrib,
    Ming Li\textsuperscript{\rm 3}\corresponding
}

\affiliations{
    \textsuperscript{\rm 1}
    Shanghai Institute for Mathematics and Interdisciplinary Sciences,
    Fudan University\\
    \textsuperscript{\rm 2}
    Shanghai Innovation Institute\\
    \textsuperscript{\rm 3}
    University of Waterloo\\
    24114020005@m.fudan.edu.cn, mli@uwaterloo.ca
}

\begin{document}

\maketitle

\begin{abstract}
Large language models (LLMs) pose challenges to academic integrity and peer review. Yet generative plagiarism detection remains an underexplored and largely unresolved challenge. Prior work on LLM-generated-text detection targets AI involvement, which may be permissible, rather than source reuse, while similarity-based methods struggle after extensive rewriting and multi-source synthesis. Motivated by the description-length view of probabilistic prediction, in which relevant side information can reduce a target sequence's code length, we introduce Source-Conditioned Description-Length Gain (SCDG), a directional, training-free framework that contrasts a frozen language model's description length of a suspicious document $P$ with and without a candidate source $S$. This contrast yields token-level log-likelihood gains that measure the incremental predictive evidence supplied by $S$.
We evaluate SCDG on the PAN at CLEF benchmarks for generative plagiarism. On a PAN 2025-derived pairwise benchmark, SCDG achieves 0.92 Precision, 0.97 Recall, and 0.94 F1, outperforming all baselines; on PAN 2026's multi-source retrieval task, it reaches 0.83 nDCG@10 and 0.96 Recall@100, surpassing all baselines. On a same-topic, same-event Multi-News test, the calibrated gain-distribution SCDG classifier predicts source reuse for only $0.125\%$ of pairs, supporting robustness to topical overlap under this evaluation protocol. These results establish SCDG as a unified and token-decomposable signal for source-specific content reuse under extensive transformation.

\end{abstract}


\section{Introduction}
\label{sec:introduction}

As LLMs evolve from writing assistants into research agents, they are becoming
embedded in scholarly knowledge production. They can retrieve sources,
synthesize evidence, organize arguments, and draft reports with limited human
intervention. These capabilities enable deeper forms of rewriting, in which
the central ideas, evidence, reasoning patterns, and organizational logic of
source materials are recombined into fluent new prose with little recognizable
lexical overlap. A document may therefore depend heavily on a particular
source while bearing limited surface resemblance to it. Detecting this form of
\emph{generative plagiarism} requires identifying source reuse under extensive
transformation.

\begin{figure}[!t]
    \centering
    \includegraphics[
        width=0.88\columnwidth,
        keepaspectratio
    ]{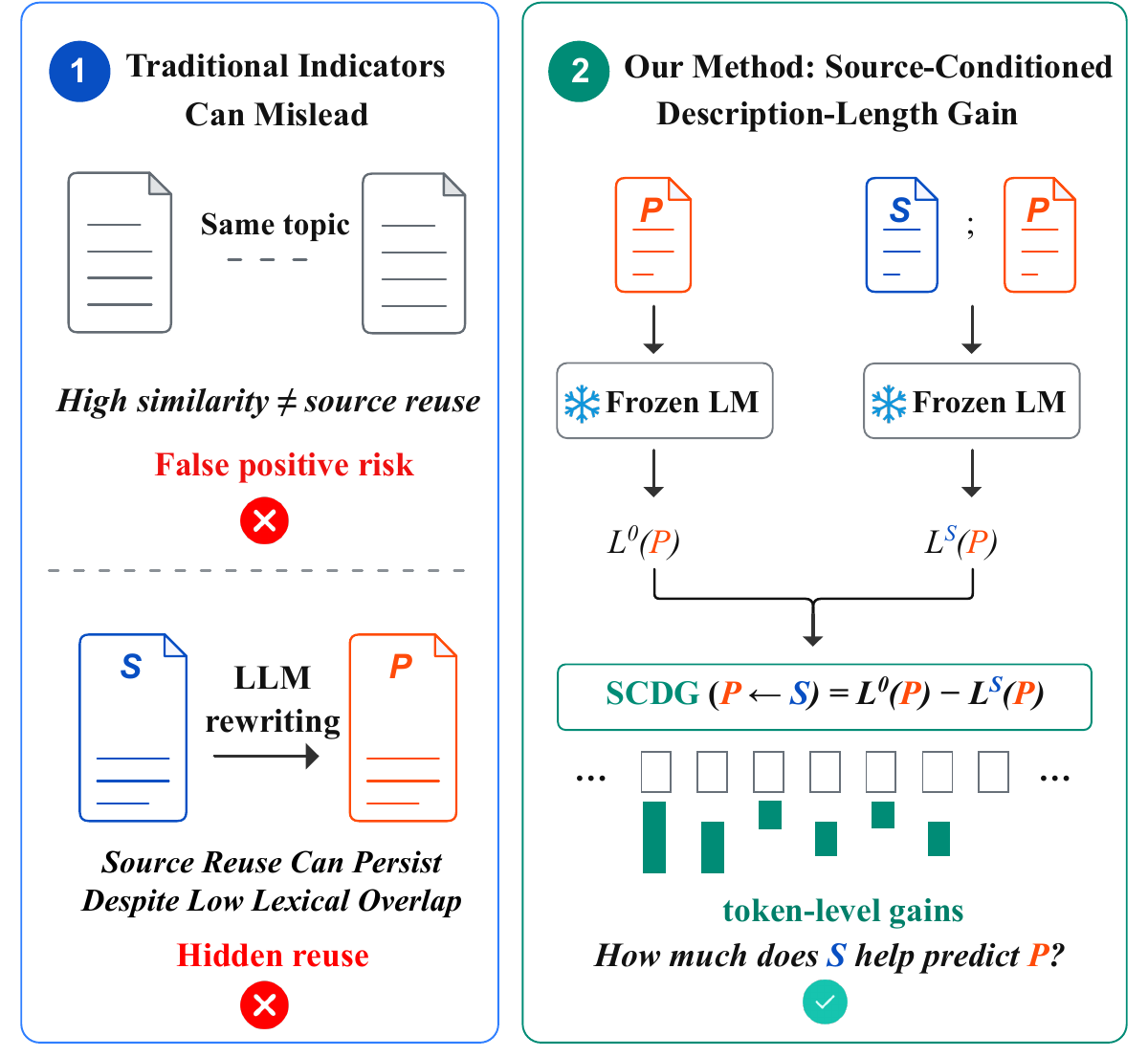}
    \caption{SCDG distinguishes source-conditioned predictive evidence
    from surface similarity under topical overlap and extensive rewriting.}
    \label{fig:intro}
\end{figure}

LLM-generated-text detection does not address the same question. Existing
classifiers and likelihood-based detectors estimate whether an LLM contributed
to a text \cite{yang2024survey,mitchell2023detectgpt,bao2024fastdetectgpt}, whereas
generative plagiarism detection asks whether the text reuses content from a particular source. AI involvement neither implies nor is required for such reuse: machine-generated text may be independently produced, while source-derived content may remain after substantial human revision.

Classical external plagiarism detection typically combines source retrieval
with passage alignment \cite{potthast2010evaluation,foltynek2019academic}. Lexical matching and BM25 work well when reused passages retain recognizable wording, while
embedding-based methods extend the comparison to semantic similarity. Both signals, however, weaken under extensive paraphrasing, reorganization, and multi-source synthesis
\cite{wahle2021neural,wahle2022transforming,greinerpetter2025overview}. Prompted LLMs can also judge document pairs directly \cite{lee2025plagbench}, but their decisions may depend on prompting and do not yield a stable measure of the contribution made by each candidate source.

The remaining difficulty is that resemblance does not reliably indicate source dependence. Independently written documents about the same topic may be highly similar, whereas a deeply rewritten document may be only weakly similar to the source on which it relies. We therefore seek a directional, source-specific signal that measures how much a candidate source $S$ contributes to a suspicious document $P$, while controlling for the baseline predictability of $P$. This motivates a different view of source evidence based on description length. Under a fixed probabilistic model, the negative log-probability of a sequence corresponds to its model-relative codelength \cite{shannon1948mathematical,rissanen1978modeling,mackay2003information}. We compare the codelength assigned to the same suspicious document with and without a candidate source in the model context. When the document reuses information from that source, conditioning on it should make the reused content easier to predict and thereby shorten the document's codelength. The resulting reduction measures the incremental predictive contribution of the candidate source. 

\begin{figure*}[t]
    \centering
    \includegraphics[width=\textwidth]{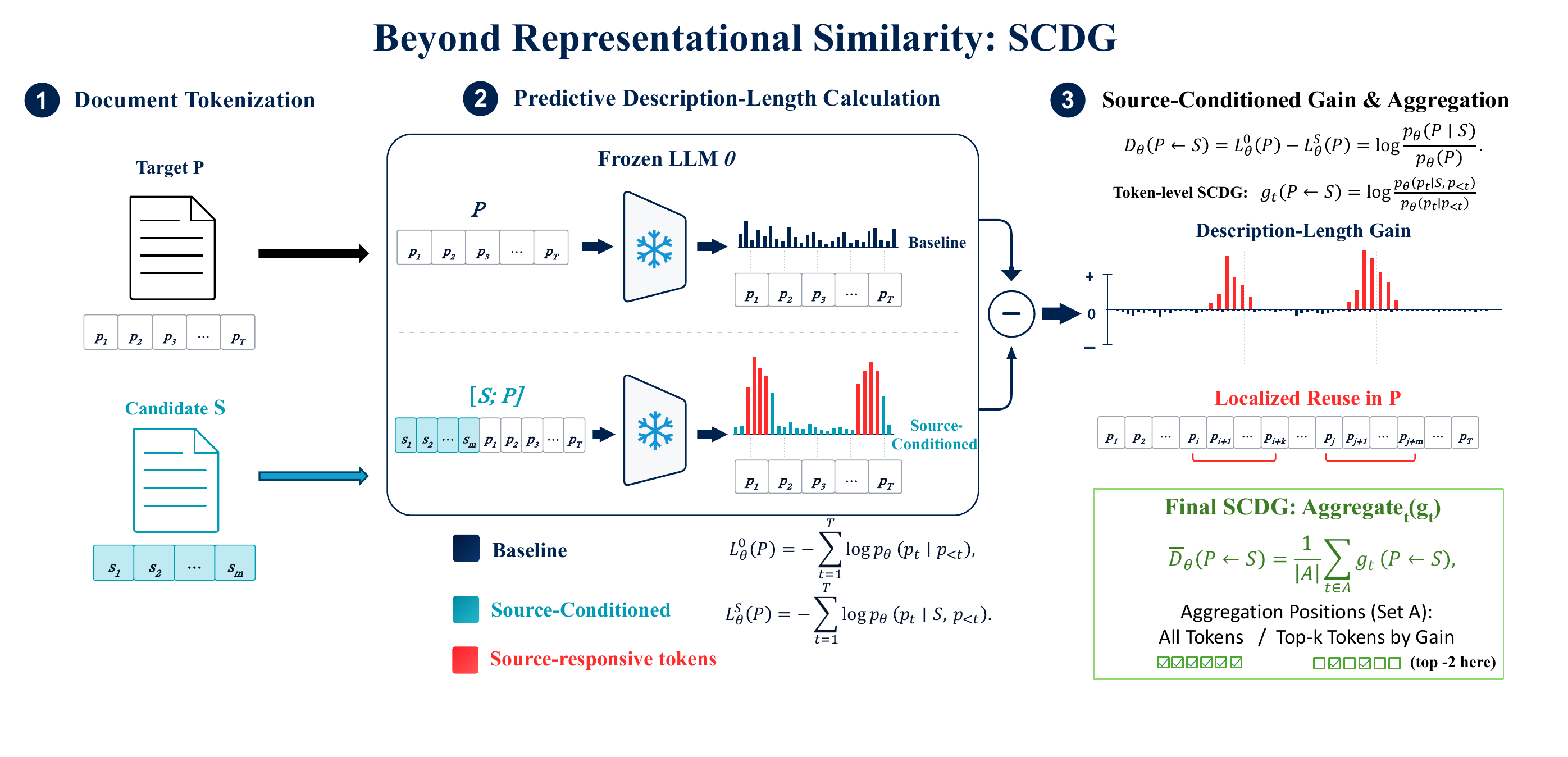}
    \caption{Overview of Source-Conditioned Description-Length Gain (SCDG).
    Given a suspicious document $P$ and a candidate source $S$, both documents
    are tokenized, and a frozen autoregressive language model scores the
    identical target-token sequence $P$ under two contexts: the target prefix
    alone and the source-conditioned context $[S;P]$. The resulting baseline
    and source-conditioned codelengths define the total description-length gain
    $D_\theta(P\leftarrow S)=L_\theta^0(P)-L_\theta^S(P)$, which decomposes
    exactly into token-level gains $g_t(P\leftarrow S)$. Positive gains identify
    source-responsive positions at which conditioning on $S$ makes the observed
    target tokens more predictable, thereby revealing localized source reuse.
    Finally, SCDG averages the gains over all target tokens or the top-$k$ tokens to obtain the length-normalized score $\overline{D}_\theta(P\leftarrow S)$ for pairwise plagiarism detection and candidate-source reranking.}
    \label{fig:main}
\end{figure*}

Building on this idea, we introduce \emph{Source-Conditioned Description-Length Gain} (SCDG), a training-free framework that measures the predictive contribution of a candidate source, as illustrated in Figure~\ref{fig:main}. SCDG compares the description
length assigned by a frozen language model to the same suspicious document with and without the source in context. A larger reduction indicates that the source provides information that helps the model predict the observed text. Because this reduction decomposes over individual target tokens, SCDG also reveals where source conditioning provides predictive evidence and allows sparse source-derived content to be aggregated without being diluted by the rest of the document. Unlike lexical or embedding similarity, SCDG measures an asymmetric relation from a candidate source to a suspicious document while controlling for the document's baseline predictability. The resulting score can be used both to classify suspicious--source pairs and to rerank candidate sources retrieved
from a large corpus, without task-specific parameter updates. SCDG should be interpreted as model-relative evidence of source dependence rather than direct proof of provenance or authorial intent.

We evaluate SCDG on both pairwise detection and full-corpus source retrieval using the PAN generative-plagiarism benchmarks. On the PAN 2025-derived pairwise benchmark, SCDG performs consistently across three frozen language models and reaches an $F_1$ score of $0.9433$, outperforming lexical, embedding-based, and prompted-LLM baselines. On PAN 2026, SCDG substantially improves two matched candidate rankings, raising nDCG@10 by more than $0.13$ without changing their candidate sets. These results show that source-conditioned predictability provides a robust signal for detecting transformed source reuse and complements conventional retrieval methods in large-scale source attribution. We further conduct two targeted analyses of the SCDG signal. A controlled source-retention study shows that SCDG increases as more genuine evidence is retained for reused content, while remaining comparatively stable on newly written content. On a complementary Multi-News test set, SCDG-based classifier produces positive decisions for only $0.125\%$ of same-topic, same-event article pairs, indicating limited susceptibility to topical confounding under this evaluation protocol. Together, these analyses examine both SCDG's sensitivity to genuine source evidence and its ability to distinguish source dependence from topical overlap.

Our main contributions are as follows:
\begin{itemize}
    \item We approach generative plagiarism detection through the lens of source-conditioned predictability and introduce \emph{Source-Conditioned Description-Length Gain} (SCDG), a training-free, token-level framework that measures the predictive contribution of a candidate source to a suspicious document.

    \item We develop DAAC, an uncertainty-aware retrieval-fusion method that
    combines lexical, sentence-level dense, and abstract-level evidence using
    retrieval ranks alone. DAAC constructs compact candidate sets without
    additional neural inference and, together with SCDG, enables scalable
    retrieval and attribution of potential source documents from large corpora.

    \item We provide a comprehensive empirical evaluation of SCDG. Our method outperforms all evaluated baselines on both the PAN 2025-derived and PAN 2026 benchmarks, while source-retention and Multi-News experiments further support its sensitivity to genuine source evidence and robustness to topical confounding.
\end{itemize}

\section{Related Work}

\paragraph{External plagiarism detection.}
Classical external plagiarism detection treats the task as source retrieval followed by passage alignment. PAN established standardized corpora and evaluation measures for this setting \cite{potthast2010evaluation}; subsequent editions separated retrieval from detailed comparison and evaluated increasingly realistic paraphrase, translation, and summarization cases \cite{potthast2011overview,potthast2012overview,potthast2013overview,potthast2014overview}. These benchmarks showed strong performance on verbatim reuse but persistent difficulty with heavily obfuscated reuse. 
Most such systems nevertheless decide reuse through explicit surface or representational similarity. This signal weakens when rewriting removes local correspondence or when evidence is distributed across multiple sources. Our method instead asks whether a candidate source makes the observed suspicious text more predictable under a fixed autoregressive model.

\paragraph{Generative plagiarism detection.}
Neural paraphrasers can maintain high semantic similarity to source texts while making paraphrase detection substantially more difficult \cite{wahle2021neural}; with larger autoregressive language models, GPT-3 paraphrases were nearly indistinguishable from original texts to human judges (53\% accuracy), while the best tested model achieved only about 66\% macro-F1 in detecting them \cite{wahle2022transforming}. A study extending language-model memorization analysis further identifies verbatim, paraphrase, and idea-level reuse from training data, but its high-precision retrieval-and-alignment pipeline yields conservative lower-bound estimates because of limited recall \cite{lee2023plagiarize}. PlagBench expands benchmark coverage with 46.5K synthetic text pairs spanning verbatim copying, paraphrasing, and summarization, and demonstrates strong plagiarism-detection performance for GPT-4 Turbo, although its advantage varies across tasks and baselines \cite{lee2025plagbench}. However, such prompted LLM judgments are sensitive to prompting and do not directly quantify the incremental contribution of a candidate source to the observed target. PAN 2025 reported promising paragraph-level performance for embedding-based systems, reaching about $0.8$ recall at roughly $0.5$ precision, but also revealed weak generalization to earlier PAN data \cite{greinerpetter2025overview}. PAN 2026 moves to a more realistic one- or multi-source generation setting, placing greater emphasis on robustness to deep rewriting, multi-source synthesis, and distribution shift \cite{bevendorff2026overview}.
\paragraph{LLM-generated-text detection.}
A separate literature detects AI authorship using supervised classifiers, zero-shot likelihood statistics, or watermarking \cite{yang2024survey}. DetectGPT obtains strong zero-shot discrimination from probability curvature \cite{mitchell2023detectgpt}, Fast-DetectGPT substantially improves its efficiency and accuracy through conditional curvature \cite{bao2024fastdetectgpt}, and ImBD targets machine-revised text by aligning to machine stylistic preferences \cite{chen2025imitate}. These methods detect LLM involvement rather than source-specific reuse, even though LLM assistance does not necessarily imply plagiarism. SCDG instead measures the model-relative predictive evidence supplied by a candidate source.

\section{Methodology}
\label{sec:methodology}


\subsection{Theoretical Foundation: Conditional Description Length}

According to Shannon’s source coding theory and its MDL interpretation, the negative log-probability of a sequence under a fixed probabilistic model corresponds to its ideal model-relative codelength; a lossless code can realize this length up to an additive coding constant \cite{shannon1948mathematical,rissanen1978modeling,mackay2003information}. Near-optimal compression of natural language requires more than modeling surface statistics or merely representing semantic content; it requires capturing the deeper generative regularities underlying linguistic structure and meaning \cite{shannon1951prediction,mahoney1999text,deletang2024language,li2025lossless}. From an algorithmic-information-theoretic perspective, the conditional codelength induced by a fixed computable model provides a computable, model-relative upper bound on conditional Kolmogorov complexity, up to an additive model-dependent constant \cite{li2008introduction}. Consequently, a substantial reduction in conditional codelength indicates that, conditioning on the candidate source yields a model-relative incremental information gain for predicting the target. This signal allows our framework to remain informative despite surface-form changes introduced by LLM paraphrasing and capture source-derived information that survives rewriting. Using natural logarithms, we define the unconditional codelength $L_\theta^0(P)$ and the source-conditioned codelength $L_\theta^S(P)$ of document $P$, measured in nats, as:
\begin{align}
L_\theta^0(P)
&= -\sum_{t=1}^{T}
\log p_\theta(p_t\mid p_{<t}),
\label{eq:unconditional-codelength}\\
L_\theta^S(P)
&= -\sum_{t=1}^{T}
\log p_\theta(p_t\mid S,p_{<t}).
\label{eq:conditional-codelength}
\end{align}

The total description-length gain yielded by the candidate source $S$ is quantified by their reduction:
\begin{equation}
D_\theta(P\leftarrow S)
= L_\theta^0(P)-L_\theta^S(P)
= \log\frac{p_\theta(P\mid S)}{p_\theta(P)}.
\label{eq:total-description-length-gain}
\end{equation}

Mathematically, $D_\theta(P\leftarrow S)$ is a model-relative
conditional log-likelihood ratio and can be viewed as a PMI-like
information-density quantity under $p_\theta$. Our contribution lies
not in this algebraic identity alone, but in its operationalization as
a directional source-reuse signal, its exact tokenwise decomposition
and sparse evidence aggregation, its unified use for pairwise
detection and candidate-source reranking, and its empirical validation
under extensive rewriting and multi-source retrieval.

In autoregressive language modeling, negative log-likelihood quantifies how surprising, and hence how difficult to predict, an observed sequence is under the model. Therefore, $D_\theta$ intuitively quantifies how much easier it is for the model to predict the observed document $P$ when guided by source $S$. When $S$ is provided in the prompt context, it can act as a probabilistic generative blueprint. If $P$ reuses the ideas, organization, or reasoning patterns of $S$, the model can exploit this source-provided generative logic rather than predict $P$ solely from its own prefix, potentially causing the codelength of $P$ to drop substantially.

\subsection{Validating the Theoretical Intuition through Controlled Source-Evidence Retention}

The preceding analysis connects conditional codelength reduction to source-guided predictability, but does not establish whether the observed gain is driven by genuinely reused source evidence. We therefore conduct a controlled source-evidence intervention on 300 aligned suspicious--source pairs from the PAN 2025 validation corpus. For each pair, we construct a source context $S^{(r)}$ retaining a proportion $r\in{0,0.25,0.50,0.75,1}$ of the annotated source evidence, replacing removed chunks with length-matched distractor passages randomly sampled from an external news corpus. Each retention level is evaluated over five random seeds under otherwise fixed scoring conditions, with length-normalized gains averaged separately for plagiarism-labeled and new content.




\begin{figure}[!t]
    \centering
    \includegraphics[width=1\linewidth]{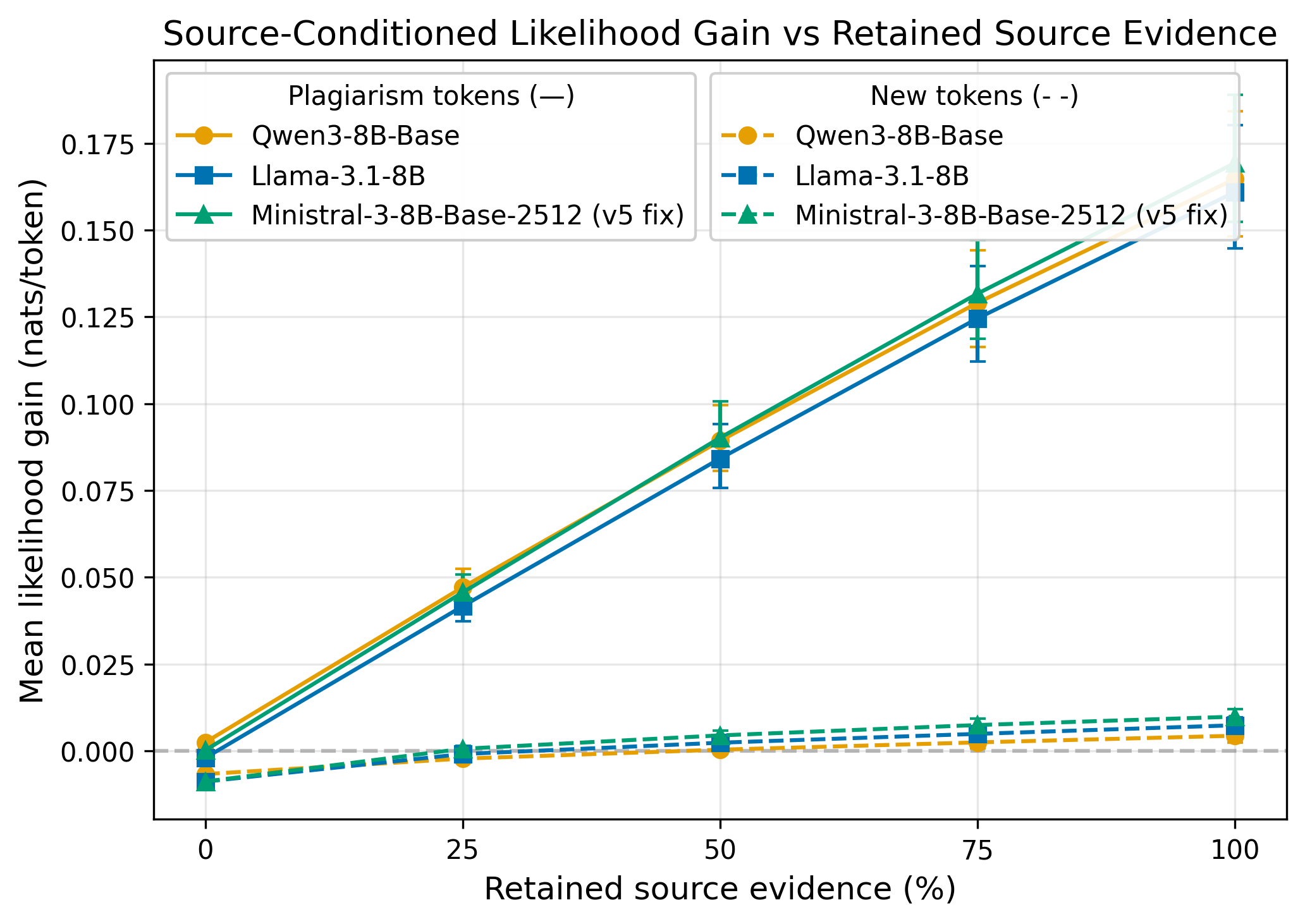}
    \caption{Mean length-normalized description-length gain over plagiarism-labeled and new content as a function of the retained source-evidence ratio. Error bars denote 95\% confidence intervals. Across all three frozen language models, plagiarism-labeled content exhibits a strong monotonic response to genuine source evidence, whereas gains for new content remain close to zero.}
    \label{fig:source-retention-intervention}
\end{figure}

As shown in Figure~\ref{fig:source-retention-intervention}, plagiarism-labeled content exhibits a strong and monotonic response to retained source evidence across Qwen3-8B-Base\cite{yang2025qwen3}, Llama-3.1-8B\cite{grattafiori2024llama3}, and Ministral-3-8B-Base-2512\cite{liu2026ministral3}. In contrast, the gain of new content remains close to zero across all retention levels and shows little response to increasing retention of genuine source evidence. This consistent separation supports our design premise: description-length gain is selectively sensitive to genuinely reused source evidence and can therefore serve as an effective signal for generative plagiarism detection.

\subsection{Source-Conditioned Description-Length Gain (SCDG)}
The pairwise codelength contrast is model-relative and does not assume that $p_\theta$ induces a universal or optimal code. Both likelihood terms are evaluated by the same frozen model over the same observed target-token sequence, differing only in whether the candidate source $S$ is absent or available as side information. 

Because autoregressive codelength is additive over target positions, the document-level gain decomposes exactly into tokenwise contributions. We define the token-level SCDG as
\begin{align}
g_t(P\leftarrow S)
&= \log\frac{p_\theta(p_t\mid S,p_{<t})}
                 {p_\theta(p_t\mid p_{<t})},
\label{eq:token-scdg}\\
D_\theta(P\leftarrow S)
&= \sum_{t=1}^{T}g_t(P\leftarrow S).
\label{eq:gain-decomposition}
\end{align}

For a selected set of target positions $A\subseteq\{1,\ldots,T\}$, the corresponding length-normalized gain is measured in nats per token. 
\begin{equation}
\overline{D}_\theta(P\leftarrow S)
=
\frac{1}{|A|}
\sum_{t\in A}g_t(P\leftarrow S).
\label{eq:normalized-scdg}
\end{equation}
This formulation supports exploratory inspection of token-level
contributions as well as document-level aggregation; the selection of $A$
used for detection is specified in the experimental setup. Specifically,
$A$ contains either all target-token positions for document-wide averaging
or the positions of the top-$k$ gains for sparse-evidence aggregation.

A positive $g_t(P\leftarrow S)$ means that conditioning on $S$ increases the probability assigned by the model to the observed token $p_t$, thereby shortening its codelength. The resulting sequence of tokenwise gains provides a directional evidence profile indicating where the candidate source improves prediction of the suspicious document.

\subsection{Pairwise Detection and Candidate-Source Reranking}

We use $\overline{D}_\theta(P\leftarrow S)$ to denote the resulting scalar score. This unified score supports both pairwise detection and candidate-source reranking without task-specific fine-tuning or parameter updates.

For pairwise plagiarism detection, we predict whether $P$ reuses content from $S$ by thresholding their directional description-length gain:
\begin{equation}
\label{eq:pairwise-decision}
\widehat{y}(P,S)
=
\mathbb{I}\!\left[
\overline{D}_\theta(P\leftarrow S)>\tau
\right].
\end{equation}

where $\tau$ is calibrated on validation data.

For multi-source retrieval and attribution, a first-stage lexical or dense retriever returns a candidate set $\mathcal{C}(P)$ of size $K$. We then rank the candidate sources in descending order of their description-length gains:
\begin{equation}
\pi_P
=
\operatorname{argsort}^{\downarrow}_{S\in\mathcal{C}(P)}
\overline{D}_\theta(P\leftarrow S).
\label{eq:candidate-source-ranking}
\end{equation}

where $\pi_P$ denotes the resulting ranked list. Candidates that yield larger source-conditioned reductions in the codelength of $P$ receive higher ranks. In this way, the same directional and token-decomposable signal supports both binary source-reuse detection and training-free candidate-source reranking under extensive generative rewriting.

\paragraph{Dense-Anchored Abstract Calibration (DAAC) indicator.}
For a fixed query $q$, let $r_j(x)$ denote the rank of candidate $x$
under route $j\in\{B,D,A\}$, corresponding to BM25, sentence-level
dense, and abstract-level retrieval. We convert each rank into
reciprocal-rank evidence:
\begin{equation}
e_j(x)
=
\frac{\mathbb{I}[r_j(x)<\infty]}
     {k+r_j(x)},
\qquad j\in\{B,D,A\},
\end{equation}

where $k=60$ and missing candidates receive zero evidence. For
readability, we write
$b(x)=e_B(x)$, $d(x)=e_D(x)$, and $a(x)=e_A(x)$.
The dense-route uncertainty and abstract-route confidence are
\begin{equation}
\begin{aligned}
U_D(x)
&=1-\min\{1,(k+1)d(x)\},\\
C_A(x)
&=\frac{\mathbb{I}[r_A(x)<\infty]}
        {\log_2(r_A(x)+1)}.
\end{aligned}
\end{equation}

The DAAC relevance indicator is
\begin{equation}
\begin{aligned}
s_{\mathrm{DAAC}}(q,x)
={}&d(x)\bigl(1+U_D(x)C_A(x)\bigr)\\
&+b(x)^2+a(x)^2.
\end{aligned}
\label{eq:daac-full}
\end{equation}

The gated interaction allows abstract confidence to reinforce uncertain
dense evidence, while the squared BM25 and abstract terms provide weak
complementary corrections without dominating the dense route.

Because DAAC depends only on retrieval ranks, it requires no
additional neural inference. From the union of the three retrieval
lists, $\mathcal{U}_q$, we retain
\begin{equation}
\mathcal{C}_q
=
\operatorname{TopK}_{x\in\mathcal{U}_q}
s_{\mathrm{DAAC}}(q,x),
\qquad K=1000,
\end{equation}
and apply the more expensive SCDG scoring only to $\mathcal{C}_q$.
After query-wise min--max normalization $\mathcal{N}_q(\cdot)$, the
final score is
\begin{equation}
\begin{aligned}
s_{\mathrm{final}}(q,x)
={}&(1-\lambda)\,
\mathcal{N}_q\!\bigl(s_{\mathrm{DAAC}}(q,x)\bigr)\\
&+\lambda\,
\mathcal{N}_q\!\bigl(s_{\mathrm{ours}}(q,x)\bigr),
\qquad x\in\mathcal{C}_q,
\end{aligned}
\label{eq:daac-rerank}
\end{equation}
where $\lambda\in[0,1]$ is selected on the development folds. Candidates
are ranked in descending order of $s_{\mathrm{final}}$.

\section{Experimental Setup}

We evaluate SCDG through three questions: \textbf{Q1} whether it provides a consistent pair-level signal across frozen language-model backends; \textbf{Q2} whether sparse aggregation improves full-corpus candidate-source ranking; and \textbf{Q3} whether the signal is confounded by same-topic relatedness.

\paragraph{Datasets.}
We evaluate SCDG in three settings: pairwise source-reuse detection, full-corpus source retrieval, and same-topic confounding. The PAN25-derived benchmark contains 6,791 scientific source--suspicious document pairs, including 4,777 positives and 2,014 negatives. We split the data by document-linked groups to prevent document overlap across partitions. PAN26 evaluates open-corpus multi-source retrieval with 200 suspicious queries, 86,822 candidate sources, and 614 graded query--source relevance judgments. To evaluate susceptibility to topical confounding, we construct 8,000 same-cluster article pairs from Multi-News and test whether a method assigns source-reuse decisions merely because two documents cover the same topic or event. We use a cluster-disjoint 6,400/1,600 train--test split.

\begin{table*}[t]
\centering
\footnotesize
\setlength{\tabcolsep}{3.0pt}
\renewcommand{\arraystretch}{1.05}

\begin{tabular*}{\linewidth}
{@{}>{\raggedright\arraybackslash}p{3.25cm}
l@{\extracolsep{\fill}}rrrrrrrr@{}}
\toprule
&
&
\multicolumn{6}{c}{\emph{Threshold-based}}
&
\multicolumn{2}{c}{\emph{Threshold-free}}
\\
\cmidrule(lr){3-8}
\cmidrule(lr){9-10}

\multicolumn{2}{@{}l}{Method}
& \multicolumn{1}{c}{Prec.}
& \multicolumn{1}{c}{Rec.}
& \multicolumn{1}{c}{$F_1$}
& \multicolumn{1}{c}{FPR$\downarrow$}
& \multicolumn{1}{c}{BAcc.}
& \multicolumn{1}{c}{MCC}
& \multicolumn{1}{c}{AUROC}
& \multicolumn{1}{c}{AP}
\\

\midrule

\multicolumn{10}{@{}l}{\textbf{Baselines}} \\

\multicolumn{2}{@{}l}{PAN12-style $n$-gram overlap}
& 0.8545 & 0.9602 & 0.9041 & 0.3861
& 0.7874 & 0.6416 & 0.8482 & 0.8842 \\

\multicolumn{2}{@{}l}{BM25 pair score}
& 0.7483 & 0.9539 & 0.8369 & 0.7587
& 0.5977 & 0.2826 & 0.7093 & 0.8237 \\

\multicolumn{2}{@{}l}{Linq-Embed-Mistral cosine}
& 0.7729 & 0.9497 & 0.8515 & 0.6597
& 0.6449 & 0.3874 & 0.7904 & 0.8813 \\

\multicolumn{2}{@{}l}{PlagBench zero-shot (vanilla)}
& 0.7032 & \textbf{1.0000} & 0.8258 & 0.9975
& 0.5012 & 0.0417 & 0.5476 & 0.7250 \\

\multicolumn{2}{@{}l}{PlagBench zero-shot (CoT)}
& 0.7032 & \textbf{1.0000} & 0.8258 & 0.9975
& 0.5012 & 0.0417 & 0.5012 & 0.7032 \\

\multicolumn{2}{@{}l}{PlagBench few-shot (vanilla)}
& 0.7027 & \textbf{1.0000} & 0.8254 & 1.0000
& 0.5000 & 0.0000 & 0.5576 & 0.7280 \\

\multicolumn{2}{@{}l}{PlagBench few-shot (CoT)}
& 0.7027 & \textbf{1.0000} & 0.8254 & 1.0000
& 0.5000 & 0.0000 & 0.5197 & 0.7111 \\

\midrule

\multicolumn{10}{@{}l}{\textbf{SCDG (Ours)}} \\

\multirow{2}{*}{\textit{SCDG \& Qwen3-8B}}
& \hspace{0.35em}Average
& 0.8914 & 0.9602 & 0.9233 & 0.2748
& 0.8419 & 0.7269 & 0.9161 & 0.9389 \\

& \hspace{0.35em}Top-$q$
& 0.9095 & 0.9696 & 0.9386 & 0.2277
& 0.8708 & 0.7831 & 0.9222 & 0.9379 \\

\addlinespace[1.5pt]

\multirow{2}{*}{\textit{SCDG \& Ministral-3-8B}}
& \hspace{0.35em}Average
& 0.9117 & 0.9686 & 0.9392 & 0.2203
& 0.8737 & 0.7863
& \underline{0.9334}
& \underline{0.9488} \\

& \hspace{0.35em}Top-$q$
& 0.9124
& \underline{0.9712}
& 0.9411
& 0.2215
& 0.8767
& 0.7914
& 0.9236
& 0.9351 \\

\addlinespace[1.5pt]

\multirow{2}{*}{\textit{SCDG \& Llama-3.1-8B}}
& \hspace{0.35em}Average
& \textbf{0.9183}
& 0.9696
& \underline{0.9433}
& \textbf{0.2042}
& \textbf{0.8835}
& \underline{0.8008}
& \textbf{0.9387}
& \textbf{0.9518} \\

& \hspace{0.35em}Top-$q$
& \underline{0.9171}
& 0.9707
& \textbf{0.9449}
& \underline{0.2067}
& \underline{0.8816}
& \textbf{0.8046}
& 0.9330
& 0.9449 \\

\bottomrule
\end{tabular*}

\caption{
Pair-level binary classification on PAN25.
Entries are per-metric medians over the same 20 hash-locked test
partitions. For each backbone, Average aggregates all eligible
token-level gains, whereas Top-$q$ aggregates the largest $q$
proportion of gains.
}
\label{tab:pair_classification}
\end{table*}

\begin{table*}[t]
\centering
\footnotesize
\setlength{\tabcolsep}{3.0pt}
\renewcommand{\arraystretch}{1.05}

\begin{tabular*}{\linewidth}
{@{}>{\raggedright\arraybackslash}p{4.50cm}
@{\extracolsep{\fill}}rrrrrrr@{}}

\toprule

Method
& \multicolumn{1}{c}{nDCG@10}
& \multicolumn{1}{c}{nDCG@100}
& \multicolumn{1}{c}{R@10}
& \multicolumn{1}{c}{R@100}
& \multicolumn{1}{c}{R@1000}
& \multicolumn{1}{c}{MRR}
& \multicolumn{1}{c}{MAP}
\\

\midrule

\multicolumn{8}{@{}l}{%
\textbf{Baselines}%
} \\[-1pt]

BM25-full
& 0.4962
& 0.5479
& 0.5188
& 0.7425
& 0.9221
& 0.7388
& 0.4175 \\

BM25-sentence
& 0.5049
& 0.5629
& 0.6038
& 0.8596
& 0.9821
& 0.7260
& 0.4328 \\

BGE-M3 Dense-full
& 0.4483
& 0.5022
& 0.5200
& 0.7563
& 0.9221
& 0.7586
& 0.4185 \\

BGE-M3 Dense-sentence
& 0.6240
& 0.6678
& 0.7321
& 0.9096
& 0.9708
& 0.8436
& 0.6071 \\

Four-way CombSUM
& 0.6764
& 0.7233
& 0.7458
& 0.9342
& 0.9838
& 0.9359
& 0.6578 \\

Linq sentence-to-chunk dense
& 0.7430
& 0.7679
& 0.8442
& 0.9500
& \underline{0.9854}
& 0.9468
& 0.7543 \\

\midrule

\multicolumn{8}{@{}l}{%
\textbf{SCDG (our method based on Top-1000)}$^\dagger$%
} \\[-1pt]

Three-Route RRF Coarse Ranking
& 0.6864
& 0.7321
& 0.7688
& 0.9600
& \textbf{0.9900}
& 0.9454
& 0.6570 \\

\quad + fixed SCDG reranking
& 0.8139
& \underline{0.8322}
& 0.8742
& 0.9550
& \textbf{0.9900}
& \textbf{0.9925}
& 0.8332 \\

DAAC coarse ranking
& 0.7678
& 0.7892
& 0.8579
& 0.9517
& \underline{0.9854}
& \underline{0.9642}
& 0.7737 \\

\quad + fixed SCDG reranking
& \textbf{0.8315}
& \textbf{0.8433}
& \textbf{0.9042}
& \textbf{0.9625}
& \underline{0.9854}
& \textbf{0.9925}
& \textbf{0.8532} \\

\bottomrule
\end{tabular*}

\caption{PAN26 full-corpus source retrieval over all 200 queries against 86,822 candidate sources. Every row is recomputed with the same qrels and evaluator; nDCG uses the original three relevance grades, while recall, MRR, and MAP use relevance $>0$. Each indented row applies the same development-selected Qwen3-8B SCDG configuration (Top-2\%, $N=1000$, $\alpha=.75$) to exactly the Top-1000 candidates in the preceding row, so Recall@1000 cannot change within a pair. Boldface and underlining denote the best and second-best distinct values, respectively, in each column; all tied values receive identical formatting.}
\label{tab:pan26_retrieval_all}
\end{table*}

\paragraph{Baselines.}
We compare SCDG with lexical, embedding, prompted-LLM, and retrieval-fusion baselines.

\noindent\textit{Pairwise generative plagiarism detection.}
For Q1, we evaluate (1) \textbf{PAN12-style character-$n$-gram matching} for normalized lexical overlap \citep{potthast2012overview}; (2) \textbf{BM25 pair scoring} for term-weighted lexical relevance \citep{robertson2009probabilistic}; (3) \textbf{Linq-Embed-Mistral cosine similarity}, adapted to document-pair scoring from the strongest post-hoc embedding baseline in the official PAN 2025 overview \citep{choi2024linq,greinerpetter2025overview}; and (4) \textbf{PlagBench-style LLM judging} with Meta-Llama-3-8B-Instruct under zero- and few-shot prompting, with and without chain-of-thought \citep{lee2025plagbench}.

\noindent\textit{Full-corpus candidate-source retrieval.}
For Q2, the six baselines comprise \textbf{BM25-full} and \textbf{BM25-sentence}, which use full-document and sentence-window query units over the same document-level index; \textbf{BGE-M3 Dense-full} and \textbf{BGE-M3 Dense-sentence} at the corresponding query granularities \citep{chen-etal-2024-m3}; \textbf{Four-way CombSUM}, which combines normalized scores from these four routes \citep{fox1994combination}; and \textbf{Linq-Embed-Mistral sentence-to-chunk retrieval}. We further evaluate SCDG reranking on the top 1,000 candidates from three fixed first-stage configurations: \textbf{equal-weight three-route RRF} \citep{cormack2009reciprocal}, and \textbf{DAAC-Full}, reporting each before and after reranking.

\paragraph{Evaluation Metrics.}
For pairwise detection, \(F_1\) is primary; we additionally report precision, recall, FPR, balanced accuracy, MCC, AUROC, and AP. For retrieval, nDCG@10 is primary, supplemented by nDCG@100, Recall@10/100/1000, MRR, and MAP. nDCG preserves the original relevance grades, whereas the remaining retrieval metrics treat relevance \(>0\) as relevant. For same-topic confounding, we report a proxy false-positive rate by operationally treating same-cluster article pairs as non-reuse cases. Lower values indicate a greater ability to distinguish source reuse from shared topic or event information.

\paragraph{Implementation Details.}
We instantiate SCDG with three frozen autoregressive backends:
Qwen3-8B-Base, Meta-Llama-3.1-8B, and
Ministral-3-8B-Base-2512. The unconditional and
source-conditioned passes score identical target-token IDs,
differing only in whether the candidate source is included in the
context. For pairwise detection, aggregation hyperparameters and
decision thresholds are selected on the training portion of each
split and fixed for test evaluation. For PAN26, SCDG reranks the
fixed Top-1,000 candidates from each first-stage system, with all
aggregation and fusion hyperparameters selected on development
queries only. Full implementation and computing-environment
details are reported in the supplement.
\section{Results and Analyses}
\subsection{Pair-Level Binary Classification}
\label{subsec:pair-results}

\paragraph{Overall performance.}
Table~\ref{tab:pair_classification} shows that SCDG provides
substantially stronger pairwise discrimination than the lexical,
embedding, and prompted-LLM baselines across all three backends.
The perfect recall of several PlagBench variants results from
predicting nearly every pair as positive, as reflected by their
near-unit FPR and near-zero MCC. SCDG instead combines high
recall with substantially stronger rejection of non-source pairs.
Performance is broadly consistent across Qwen, Ministral, and
Llama, with Llama providing the strongest overall point
estimates, indicating that the source-conditioned signal is not
specific to a single model family.

\paragraph{Effect of sparse aggregation.}
Top-\(q\) aggregation yields slightly higher point-estimate \(F_1\) across all three backends, consistent with source evidence being localized within only part of a suspicious document.  When source-derived content
occupies only a limited portion of a suspicious document, averaging over all target tokens mixes strong positive gains on reused passages with weak, zero, or negative gains on unrelated content, thereby
diluting the document-level dependency signal. Top-$q$ mitigates this effect by emphasizing the upper tail of the token-level gain distribution, where localized source evidence is more likely to concentrate. This interpretation is also consistent with the controlled
source-retention results, which show stronger gain responses on plagiarism-labeled content than on newly written content. However, the improvement is metric-dependent: Top-\(q\) favors \(F_1\) and MCC, whereas average aggregation retains stronger precision, FPR, AUROC, and AP for the Llama backend. Sparse aggregation therefore changes the operating characteristics of SCDG rather than uniformly improving its discrimination, making the preferred aggregation dependent on the application objective.

\subsection{Full-Corpus Source Retrieval}



\paragraph{Retrieval granularity and fusion.}
Table~\ref{tab:pan26_retrieval_all} shows that retrieval granularity strongly affects candidate quality. Sentence-level dense retrieval substantially outperforms full-document dense retrieval at both the head and deeper cutoffs, whereas sentence segmentation primarily improves the deeper recall of BM25. Linq sentence-to-chunk retrieval is the strongest single-route baseline, and conventional CombSUM does not surpass it at the head of the ranking. This indicates that additional retrieval routes are useful only when their relative reliability is properly controlled.



\paragraph{Effect of SCDG reranking.}
Across both matched candidate sets, SCDG improves nDCG@10 and
MAP without changing Recall@1000, showing that its gains arise
from reordering rather than additional retrieval. The larger
improvement over RRF suggests that SCDG complements a strong
first-stage retriever, with most gains concentrated near the
ranking head. This promotion of stronger contributing sources
can move weaker relevant candidates across deeper cutoffs,
explaining the slight Recall@100 decrease for RRF. As MRR is
nearly saturated, nDCG and MAP are more informative for
evaluating multi-source ordering. DAAC with SCDG yields the
strongest displayed endpoint, whereas the matched RRF comparison
more clearly isolates the contribution of reranking.


\begin{table}[t]
\centering
\footnotesize
\setlength{\tabcolsep}{4pt}
\renewcommand{\arraystretch}{1.02}

\begin{tabular}{@{}p{4.55cm}r@{}}
\toprule
Method & \multicolumn{1}{c}{Proxy FPR (\%) $\downarrow$} \\
\midrule

\multicolumn{2}{@{}l}{\textit{Lexical and retrieval baselines}} \\[-1.5pt]
\hspace{0.8em}BM25 pair score
& 71.1250\% \\
\hspace{0.8em}Max 4-gram containment
& 31.5625\% \\
\hspace{0.8em}4-gram logistic
& 29.9375\% \\

\addlinespace[2pt]
\multicolumn{2}{@{}l}{\textit{SCDG statistics and auxiliary calibrators}} \\[-1.5pt]
\hspace{0.8em}Average SCDG
& 62.8760\% \\
\hspace{0.8em}Positive-gain rate SCDG
& \underline{8.6875\%} \\
\hspace{0.8em}Gain-distribution logistic
& \textbf{0.1250\%} \\

\bottomrule
\end{tabular}

\caption{Proxy false-positive rates on the cluster-disjoint Multi-News same-topic stress set. }
\label{tab:pan25_multinews}
\end{table}

\subsection{Robustness to Same-Topic Confounding}
\label{subsec:same-topic-robustness}

Table~\ref{tab:pan25_multinews} reports proxy false-positive rates on the Multi-News same-topic stress set, where lower values indicate stronger resistance to topical confounding. BM25, lexical baselines, and Average SCDG remain highly sensitive to shared topic or event information, with false-positive rates of 29.9375--71.1250\%. In contrast, Positive-gain-rate SCDG reduces the rate to 8.6875\%, while the gain-distribution logistic variant reaches 0.1250\%. These results show that topical robustness depends on how the token-level gain distribution is summarized and calibrated, rather than on the mean gain alone. The
corresponding PAN25 results in the supplementary materials characterize the trade-off between detection utility and same-topic robustness.

\section{Conclusions}

In this work, we introduced SCDG, a directional and training-free framework that measures the incremental predictive evidence supplied by a candidate source through description-length contrast. Controlled interventions support its sensitivity to retained source evidence, while experiments demonstrate strong performance on PAN25 pairwise detection and PAN26 candidate-source ranking across frozen language-model backends. The Multi-News test further indicates robustness to same-topic and same-event confounding. Nevertheless, SCDG is more computationally expensive than conventional similarity measures and has been evaluated mainly on English scientific documents and a constructed news-domain stress set. Future work should improve scoring efficiency, broaden multilingual and cross-domain evaluation, localize passage-level evidence, and study more complex source reuse, while leveraging its token-level decomposition for interpretable evidence tracing. Combined with high-recall retrieval, SCDG enables scalable source attribution without task-specific model updates, supporting more transparent academic-integrity screening and source verification in scholarly writing.

\appendix











\section{Appendix}
This appendix provides additional theoretical, experimental, and implementation details of our main paper. These sections supplements the main paper with derivations, dataset and split construction, implementation details, hyperparameter settings, computing infrastructure, and extended results.

\section{Additional Theory and Method Details}
\label{sec:supp-theory-method}

\subsection{Scope and Status of the Claims}
\label{subsec:supp-claim-status}

Table~\ref{tab:supp-claim-status} distinguishes the exact identities
underlying SCDG from its coding interpretation, task-specific design
hypotheses and aggregation choices, and empirically scoped claims.

\begin{table*}[!t]
\centering
\small
\begin{tabularx}{\textwidth}{@{}p{2.2cm}XX@{}}
\toprule
Status & Statement & Scope or supporting evidence \\
\midrule
Exact identity
& $D_\theta(P\leftarrow S)
=\log p_\theta(P\mid S)-\log p_\theta(P)$.
& Follows from the definitions of the two ideal model-relative
codelengths. \\

Exact identity
& $D_\theta(P\leftarrow S)=\sum_t g_t(P\leftarrow S)$.
& Follows from autoregressive factorization over the identical target
token sequence. \\

Coding interpretation
& $-\log p_\theta$ is an ideal codelength under the fixed probabilistic
model.
& Model-relative; it does not require or imply that $p_\theta$ is a
universal or optimal code
\citep{shannon1948mathematical,rissanen1978modeling,mackay2003information}. \\

Design hypothesis
& Retaining more annotated source evidence should produce a stronger
gain response for plagiarism-labeled than for newly written content.
& Evaluated by the controlled retention experiment and its
content-type-by-retention interaction. \\

Aggregation choice
& Upper-tail aggregation can reduce dilution when source-responsive
positions are sparse.
& Evaluated against document-wide averaging on PAN25 and through
candidate-source reranking on PAN26; it is not uniformly superior for
every metric. \\

Empirical scope
& SCDG supports pairwise detection and candidate-source reranking.
& Restricted to the reported PAN25-derived and PAN26 evaluations and
the tested frozen language-model backends; the Multi-News result is a
proxy rate under reuse-unknown same-topic pairs. \\
\bottomrule
\end{tabularx}
\caption{Status of the principal theoretical, methodological, and
empirical statements. ``Exact identity'' denotes an algebraic
consequence of the definitions; the remaining rows distinguish an
established coding interpretation, task-specific design choices and
hypotheses, and empirically scoped claims.}
\label{tab:supp-claim-status}
\end{table*}

\subsection{Model-Relative Codelength}
\label{subsec:supp-model-relative-codelength}

Let $P=(p_1,\ldots,p_T)$ be the target-token sequence of the
suspicious document, $S$ a candidate source, and $p_\theta$ a fixed
autoregressive language model. Autoregressive factorization gives
\begin{align}
p_\theta(P)
&=\prod_{t=1}^{T}p_\theta(p_t\mid p_{<t}),
\label{eq:supp-autoregressive-unconditional}\\
p_\theta(P\mid S)
&=\prod_{t=1}^{T}p_\theta(p_t\mid S,p_{<t}).
\label{eq:supp-autoregressive-conditional}
\end{align}
Using natural logarithms, the corresponding ideal model-relative
codelengths, in nats, are
\begin{align}
L_\theta^0(P)
&=-\log p_\theta(P)
=-\sum_{t=1}^{T}\log p_\theta(p_t\mid p_{<t}),
\label{eq:supp-unconditional-codelength}\\
L_\theta^S(P)
&=-\log p_\theta(P\mid S)
=-\sum_{t=1}^{T}\log p_\theta(p_t\mid S,p_{<t}).
\label{eq:supp-conditional-codelength}
\end{align}
The term ``ideal'' is important: source coding associates
$-\log p$ with an ideal code length, while an implementable
integer-length lossless code may differ by a bounded coding overhead
\citep{shannon1948mathematical,mackay2003information}. Throughout the
paper, both quantities are evaluated directly from the same frozen
model and therefore remain model-relative.

From an algorithmic-information perspective, a fixed computable
probabilistic model induces a computable code. Expressing lengths in
bits, this yields a model-dependent upper bound of the form
\begin{equation}
K(P\mid S)
\leq
K(\theta)
+\frac{L_\theta^S(P)}{\ln 2}
+O(1),
\label{eq:supp-kolmogorov-upper-bound}
\end{equation}
where the model-description and coding terms are independent of the
particular target once the coding scheme is fixed. This relationship
motivates the conditional-description-length interpretation, but SCDG
does not claim to compute conditional Kolmogorov complexity
\citep{li2008introduction}.

\subsection{From Codelength Reduction to SCDG}
\label{subsec:supp-scdg-derivation}

Subtracting the source-conditioned codelength from the unconditional
codelength gives
\begin{align}
D_\theta(P\leftarrow S)
&=L_\theta^0(P)-L_\theta^S(P)
\nonumber\\
&=-\log p_\theta(P)+\log p_\theta(P\mid S)
\nonumber\\
&=\log\frac{p_\theta(P\mid S)}{p_\theta(P)}.
\label{eq:supp-total-scdg}
\end{align}
Thus, $D_\theta(P\leftarrow S)$ is exactly a conditional
log-likelihood ratio under the fixed model. It compares two
probabilities of the same observed target sequence and differs only in
whether $S$ is available as side information.

If $p_\theta(P\mid S)$ and $p_\theta(P)$ are viewed as compatible
conditional and marginal components of a joint probabilistic model,
Equation~\ref{eq:supp-total-scdg} has the same algebraic form as the
pointwise information density
\begin{equation}
i_\theta(P;S)
=
\log\frac{p_\theta(P,S)}
{p_\theta(P)p_\theta(S)}
=
\log\frac{p_\theta(P\mid S)}{p_\theta(P)}.
\label{eq:supp-information-density}
\end{equation}
We therefore describe SCDG as closely related to model-relative
information density, or as a PMI-like quantity. We do not interpret it
as an estimate of population mutual information: no expectation over a
joint distribution of $(P,S)$ is taken, and the implemented
probabilities are obtained from two prompted contexts of the same
frozen language model.

The score is directional. In general,
\begin{equation}
D_\theta(P\leftarrow S)
\neq
D_\theta(S\leftarrow P),
\label{eq:supp-directionality}
\end{equation}
because the target sequence, its autoregressive prefix, and the
conditioning context are exchanged. This directionality is useful for
the source-reuse task, which asks how much a candidate source improves
prediction of the suspicious document rather than whether the two
documents are symmetrically similar.

\subsection{Exact Tokenwise Decomposition}
\label{subsec:supp-token-decomposition}

Substituting Equations~\ref{eq:supp-unconditional-codelength}
and~\ref{eq:supp-conditional-codelength} into
Equation~\ref{eq:supp-total-scdg} yields
\begin{align}
D_\theta(P\leftarrow S)
&=\sum_{t=1}^{T}\log p_\theta(p_t\mid S,p_{<t})\nonumber\\[-1mm]
&\quad-\sum_{t=1}^{T}\log p_\theta(p_t\mid p_{<t}).
\label{eq:supp-token-decomposition}
\end{align}
Defining
\begin{equation}
\begin{aligned}
g_t(P\leftarrow S)={}&\log p_\theta(p_t\mid S,p_{<t})\\[-1mm]
&-\log p_\theta(p_t\mid p_{<t})
\end{aligned}
\label{eq:supp-token-gain}
\end{equation}
therefore gives the exact identity
\begin{equation}
D_\theta(P\leftarrow S)
=
\sum_{t=1}^{T}g_t(P\leftarrow S).
\label{eq:supp-additivity}
\end{equation}
A positive $g_t$ means that the source-conditioned context assigns
greater probability to the observed target token than the
unconditional context. This is a source-responsive predictive
contribution at position $t$; without a separate localization
evaluation, it should not be interpreted as a verified passage-level
reuse label.

For a selected set of eligible target positions
$A\subseteq\{1,\ldots,T\}$, the normalized aggregation is
\begin{equation}
\overline{D}_{\theta,A}(P\leftarrow S)
=
\frac{1}{|A|}
\sum_{t\in A}g_t(P\leftarrow S).
\label{eq:supp-normalized-gain}
\end{equation}
When $A$ contains every eligible target position,
Equation~\ref{eq:supp-normalized-gain} is the document-wide mean.
For sparse aggregation, $A$ contains the configured number or
proportion of eligible positions with the largest gains. The full sum
in Equation~\ref{eq:supp-additivity} remains an exact codelength
difference; selecting the upper tail and averaging it is a
task-specific aggregation rule designed to prevent sparse positive
evidence from being diluted by many weak, zero, or negative
contributions.

\subsection{Assumptions and Limitations of the Signal}
\label{subsec:supp-assumptions}

The interpretation of SCDG relies on the following explicit
conditions and restrictions.

\begin{enumerate}
    \item \textbf{Fixed-model comparison.}
    Both likelihoods must be produced by the same frozen language
    model and tokenizer.

    \item \textbf{Identical target support.}
    Both passes must score the identical target-token sequence. If the
    target tokenization or evaluated positions differ, the
    codelengths are not directly comparable.

    \item \textbf{Conditioning-only contrast.}
    The intended difference between the two passes is whether $S$ is
    present as side information. Prompt formatting, truncation, and
    positional effects can also influence the model and must therefore
    be held fixed as far as the implementation permits.

    \item \textbf{Model-relative evidence.}
    The score reflects the predictive behavior of $p_\theta$, not a
    universal description length or a model-independent measure of
    dependence.

    \item \textbf{No provenance guarantee.}
    Positive gain does not by itself establish that $P$ was historically
    derived from $S$. Shared topic, entities, terminology, style, or
    discourse organization may also improve prediction.

    \item \textbf{Context availability.}
    The candidate source and the evaluated target positions must fit
    the implemented context and truncation policy. Evidence outside
    the retained context cannot affect the score.

    \item \textbf{Aggregation dependence.}
    Document-wide and upper-tail aggregation have different operating
    characteristics. Upper-tail selection is useful when evidence is
    sparse, but it is not itself a likelihood ratio for the complete
    document and need not improve every evaluation metric.

    \item \textbf{No validated localization claim.}
    Tokenwise decomposition is faithful to the document-level score
    and supports exploratory inspection, but the present experiments
    do not establish token- or passage-level localization accuracy.
\end{enumerate}

\subsection{Connection Between Claims and Experiments}
\label{subsec:supp-claim-experiment-map}

The experiments address different parts of the framework and should
not be interpreted interchangeably.
\begin{table*}[!t]
\centering
\small
\begin{tabularx}{\textwidth}{@{}p{3.2cm}XX@{}}
\toprule
Evaluation & Question addressed & Interpretation boundary \\
\midrule
Controlled source-evidence retention
& Whether gains over plagiarism-labeled content respond more strongly
than gains over newly written content as more annotated source
evidence is retained.
& Supports the design premise that description-length gain is
selectively sensitive to genuinely reused source evidence under the
controlled retention protocol. \\

PAN25-derived pairwise evaluation
& Whether aggregated SCDG separates labeled source-reuse and non-reuse
pairs across frozen language-model backends.
& Supports pairwise discrimination under the PAN25-derived protocol,
not universal detection across domains and languages. \\

PAN26 full-corpus retrieval
& Whether SCDG improves the ordering of fixed first-stage candidate
sets in a multi-source retrieval task.
& Demonstrates reranking utility but cannot recover relevant sources
that are absent from the first-stage candidate set. \\

Multi-News same-topic stress test
& Whether calibrated SCDG-based summaries produce positive
source-reuse decisions for documents sharing a topic or event.
& Uses reuse-unknown same-cluster pairs and therefore reports a proxy
false-positive rate under this operational protocol. \\
\bottomrule
\end{tabularx}
\caption{Relationship between the framework's empirical hypotheses
and the reported evaluations.}
\label{tab:supp-claim-experiment-map}
\end{table*}

\section{Data Appendix}
\label{sec:supp-data}

\subsection{Dataset Selection}
\label{subsec:supp-dataset-selection}

We use three datasets because they test complementary requirements of
source-reuse detection. PAN25 provides source--suspicious document
pairs with span-level XML annotations and therefore supports controlled
pairwise evaluation under generative rewriting
\citep{greinerpetter2025pan25dataset}. PAN26 changes the problem from
pair scoring to open-corpus retrieval: each suspicious scientific
document must be matched to all of its contributing sources in a large
candidate collection \citep{greinerpetter2026pan26dataset}. Multi-News
groups multiple news articles that describe the same
topic or event \citep{fabbri-etal-2019-multi}; we use these groups to
construct a deliberately difficult stress test for topical
confounding. Together, the three settings evaluate pairwise
discrimination, multi-source ranking, and susceptibility to same-topic
confounding.

\subsection{PAN25-Derived Pair Benchmark}
\label{subsec:supp-pan25-pairs}

\paragraph{Source data and document-level labels.}
The benchmark is derived from the PAN25 generative-plagiarism
validation corpus. The original task provides a list of
suspicious--source pairs, the corresponding documents, and an XML
annotation file for each pair. A \texttt{plagiarism} feature records an
aligned reused span and identifies its associated source document; an
\texttt{altered} feature identifies source-independent paraphrased
content and is not treated as plagiarism.

For every listed pair $(P,S)$, we assign a positive document-level
label if its XML file contains at least one
\texttt{feature name="plagiarism"} whose
\texttt{source\_reference} matches $S$. Otherwise, the pair receives a
negative document-level label. Thus, a negative pair means that the
PAN25 annotation contains no plagiarism relation for that particular
pair; it may still contain \texttt{altered} spans that were generated
independently of the candidate source.

\paragraph{Length filtering and resulting manifest.}
The construction script records the token counts of the suspicious and
source documents using the Qwen3-8B-Base tokenizer. Pairs whose
recorded total length exceeds 128,000 tokens are excluded. The resulting
fixed manifest is shared by all evaluated language-model backends, so
the set of evaluated pairs does not change with the scoring model.
Table~\ref{tab:supp-pan25-pair-statistics} reports its composition.

\begin{table}[!t]
\centering
\begin{tabular}{@{}lr@{}}
\toprule
Quantity & Count \\
\midrule
All pairs & 6,791 \\
Positive pairs & 4,777 \\
Negative pairs & 2,014 \\
Unique suspicious files & 6,770 \\
Unique source files & 6,768 \\
\bottomrule
\end{tabular}
\caption{Statistics of the PAN25-derived pair manifest.}
\label{tab:supp-pan25-pair-statistics}
\end{table}

Each manifest record stores
\texttt{pair\_id}, \texttt{susp\_file}, \texttt{src\_file},
\texttt{xml\_file}, \texttt{xml\_exists}, \texttt{doc\_label},
\texttt{num\_annotations\_for\_pair}, \texttt{annotations},
\texttt{susp\_tokens}, \texttt{src\_tokens}, and
\texttt{total\_tokens}. The manifest contains identifiers,
annotations, labels, and counts; the released version does not contain
the copyrighted PAN document text.

\subsection{PAN25 Document-Linked Splits}
\label{subsec:supp-pan25-splits}

\paragraph{Connected-component grouping.}
Splitting individual rows would permit the same document to occur in
both training and test data whenever it participates in more than one
pair. To prevent this leakage, we construct a bipartite graph whose
typed vertices are suspicious-file and source-file identifiers. Every
manifest pair introduces an edge between its suspicious and source
vertices. A union--find procedure computes the graph's connected
components, and all pairs in the same component are assigned to the
same partition.

\paragraph{Twenty seed-locked repetitions.}
The experiment uses 20 fixed, seed-locked train--test splits rather
than hash-locked splits. Connected components are stratified by their
numbers of positive and negative pairs. Within each stratum, component
identifiers are shuffled with
\texttt{numpy.random.default\_rng(seed)}, after which 20\% are assigned
to test and the remainder to training. The seeds are the consecutive
integers
\[
20260714,20260715,\ldots,20260733.
\]
Every repetition has the class distribution shown in
Table~\ref{tab:supp-pan25-split-statistics}.

\begin{table}[!t]
\centering
\begin{tabular}{@{}lrrr@{}}
\toprule
Partition & Positive & Negative & Total \\
\midrule
Training & 3,822 & 1,610 & 5,432 \\
Test & 955 & 404 & 1,359 \\
\midrule
All & 4,777 & 2,014 & 6,791 \\
\bottomrule
\end{tabular}
\caption{Class distribution in each of the 20 PAN25 splits.}
\label{tab:supp-pan25-split-statistics}
\end{table}

\paragraph{Leakage audit.}
For every repetition, the split index records
\texttt{suspicious\_overlap\_count=0},
\texttt{source\_overlap\_count=0}, and
\texttt{component\_overlap\_count=0}. The split-generation code also
asserts that the sets of training and test component identifiers are
disjoint. Consequently, no suspicious-file identifier, source-file
identifier, or document-linked component is shared across the two
partitions.

\subsection{PAN26 Full-Corpus Retrieval Benchmark}
\label{subsec:supp-pan26}

\paragraph{Corpus and relevance judgments.}
Each PAN26 suspicious scientific document was generated from multiple
contributing sources, and the task is to retrieve all contributing
documents from the complete candidate collection. The
\texttt{qrels.txt} file provides graded query--source relevance
judgments in TREC format; Table~\ref{tab:supp-pan26-statistics}
summarizes the corpus and label counts.

\begin{table}[!t]
\centering
\begin{tabular}{@{}lr@{}}
\toprule
Quantity & Count \\
\midrule
Suspicious queries & 200 \\
Candidate source documents & 86,822 \\
All relevance judgments & 614 \\
Grade-1 judgments & 400 \\
Grade-2 judgments & 168 \\
Grade-3 judgments & 46 \\
\bottomrule
\end{tabular}
\caption{Statistics of the PAN26 retrieval benchmark.}
\label{tab:supp-pan26-statistics}
\end{table}

For nDCG, a judgment of grade $r$ is converted to gain
$2^r-1$, preserving the three relevance levels. Recall treats every
judgment with $r>0$ as relevant. Retrieval is
evaluated against all 86,822 candidate documents. An auxiliary
4,614-row pair-format manifest is retained for derived pair-level
auditing, but it is not used as the candidate pool for the reported
full-corpus retrieval results.

\paragraph{Fixed query folds.}
The 200 query identifiers are assigned to five fixed folds recorded in
\texttt{folds.jsonl}. Their sizes and roles are shown in
Table~\ref{tab:supp-pan26-folds}. Fold 0 is the frozen 40-query
holdout, whereas Folds 1--4 form the 160-query development set used
for Stage-2 aggregation and reranking configuration selection. The
corpus-wide descriptive profile in the main paper applies one
development-selected configuration unchanged to all 200 queries.

\begin{table}[!t]
\centering
\begin{tabular}{@{}lrl@{}}
\toprule
Fold & Queries & Role \\
\midrule
0 & 40 & Frozen holdout \\
1 & 40 & Development \\
2 & 41 & Development \\
3 & 39 & Development \\
4 & 40 & Development \\
\midrule
All & 200 & \\
\bottomrule
\end{tabular}
\caption{Fixed PAN26 query-fold assignment.}
\label{tab:supp-pan26-folds}
\end{table}

\subsection{Multi-News Same-Topic Stress Test}
\label{subsec:supp-multinews}

\paragraph{Pair construction.}
The stress set is derived from the source-document side of the
Multi-News training split. In Multi-News, each record groups multiple
news articles associated with the same reference summary. We separate
the source record into its constituent, non-summary articles using the
dataset delimiter. Articles shorter than 200 characters or longer than
32,000 characters are excluded, and a cluster remains eligible only
if at least two articles survive. Candidate articles are then sampled
within the same cluster to form a pair. Construction uses
\texttt{random.seed(42)} and targets 8,000 pairs. Although the
construction code permits at most ten pairs per topic, the realized
manifest contains 8,000 distinct cluster identifiers and therefore
one selected pair per cluster.

The resulting pairs are deliberately difficult same-topic stress
cases: both articles concern the same topic or event, but neither is
the Multi-News reference summary. For operational evaluation, the
manifest assigns \texttt{doc\_label=0} to all 8,000 pairs; this label
does not constitute a verified non-reuse annotation.

\paragraph{Cluster-disjoint partition.}
We sort the unique cluster identifiers and shuffle them using
\texttt{numpy.random.default\_rng(20260714)}, then divide them as shown
in Table~\ref{tab:supp-multinews-split}. The split audit reports both
\texttt{cluster\_overlap\_count=0} and
\texttt{pair\_overlap\_count=0}; hence no cluster or selected pair
occurs in both partitions.

\begin{table}[!t]
\centering
\begin{tabular}{@{}lrr@{}}
\toprule
Partition & Pairs & Clusters \\
\midrule
Training & 6,400 & 6,400 \\
Test & 1,600 & 1,600 \\
\midrule
All & 8,000 & 8,000 \\
\bottomrule
\end{tabular}
\caption{Cluster-disjoint Multi-News partition.}
\label{tab:supp-multinews-split}
\end{table}

\paragraph{Why the reported FPR is a proxy.}
Multi-News supplies topical clusters and reference summaries, but it
does not provide provenance annotations establishing whether one
constituent news article reused another. Accordingly, the same-cluster
pairs are reuse-unknown rather than manually verified non-reuse
examples. We operationally treat them as negatives and report the
proportion predicted as source reuse. We call this quantity a
\emph{proxy false-positive rate}: an apparent positive may reflect
topical confounding, but it may also reflect an unannotated
source-reuse relation. The stress test therefore measures
susceptibility to same-topic and same-event confounding under this
protocol; it does not establish a population false-positive rate over
verified non-reuse pairs.

\subsection{Released Manifests and Construction Artifacts}
\label{subsec:supp-data-artifacts}

Table~\ref{tab:supp-data-artifacts} lists the identifier-level
artifacts needed to reconstruct the derived benchmarks and their
splits. The PAN25 pair manifest stores labels, aligned annotations,
and token counts; its split files record fixed memberships and leakage
audits. The PAN26 files record the fixed query folds and the auxiliary
pair-format identifiers. The Multi-News files record the constructed
same-cluster pairs and their cluster-disjoint partition. Filenames are
reported without machine-specific paths.

\begin{table}[!t]
\centering
\small
\begin{tabularx}{\columnwidth}{@{}p{1.25cm}Xr@{}}
\toprule
Dataset & Artifact & Rows/files \\
\midrule
PAN25
& \path{pan25val_filtered_normal_manifest_128k_6791.jsonl}
& 6,791 \\

PAN25
& Twenty split CSV files and \path{manifest_index.json}
& 20+1 \\

PAN26
& \path{folds.jsonl}
& 200 \\

PAN26
& \path{pan2026_pairs_manifest.jsonl}
& 4,614 \\

Multi-News
& \path{manifest.jsonl}
& 8,000 \\

Multi-News
& \path{multinews_split_manifest_seed20260714.csv}
& 8,000 \\
\bottomrule
\end{tabularx}
\caption{Derived manifests and split indexes supplied for
reproducibility.}
\label{tab:supp-data-artifacts}
\end{table}

The corresponding construction scripts are
\path{build_pairs_manifest.py}, \path{split_core.py},
\path{build_multi_news_pairs.py}, and
\path{prepare_multinews_split.py}.

\subsection{Access, Licensing, and Redistribution}
\label{subsec:supp-data-licensing}

PAN25 and PAN26 are accessed through their official PAN task pages:
\begin{itemize}
    \item PAN25:
    \url{https://pan.webis.de/clef25/pan25-web/}
    \item PAN26:
    \url{https://pan.webis.de/clef26/pan26-web/}
\end{itemize}
Both PAN task pages state that the datasets contain copyrighted
material, may be used only for research purposes, and may not be
redistributed.

Multi-News is obtained from its official project repository:
\url{https://github.com/Alex-Fabbri/Multi-News}.
The repository's \texttt{LICENSE.txt} is a custom LILY LAB Dataset
Usage Agreement, not the Apache 2.0 License. It permits use for
non-commercial research and educational purposes subject to the
agreement's conditions and does not grant a general license to the
underlying intellectual property.

We release no raw document text, only derived identifier-level
manifests and construction or audit scripts; users must obtain the
original datasets from their authorized sources.

\section{Complete SCDG Implementation Details}
\label{sec:supp-scdg-implementation}

This appendix specifies the implementation used to produce the SCDG
results in the main paper.  The PAN25 pairwise experiment and the PAN26
retrieval experiment use the same directional contrast and the same
target-token alignment requirement, but they use different input-length
handling.  We therefore describe the two realizations separately.  No
language-model parameter is trained or updated in either setting.

\subsection{Frozen Backends, Checkpoints, and Tokenizers}
\label{subsec:supp-scdg-backends}

All three backends are local checkpoint snapshots.  No Hugging Face
revision or commit hash was specified when the experiments were run.
For each backend, the tokenizer is loaded from the same local snapshot as
the model by \texttt{AutoTokenizer.from\_pretrained}; there is no
independent tokenizer revision.  To make the local snapshots auditable,
the run protocol records SHA256 fingerprints for the available model and
tokenizer metadata files, including \texttt{config.json},
\texttt{tokenizer\_config.json}, \texttt{tokenizer.json},
\texttt{tokenizer.model}, \texttt{tekken.json}, and
\texttt{model.safetensors.index.json} when present.

\begin{table*}[!t]
\centering
\scriptsize
\setlength{\tabcolsep}{3pt}
\renewcommand{\arraystretch}{1.06}
\begin{tabularx}{\textwidth}{@{}p{2.25cm}p{2.6cm}p{2.3cm}p{2.2cm}X@{}}
\toprule
Backend & Local snapshot & Model loader & Tokenizer & Runtime configuration \\
\midrule
Qwen3-8B-Base
& \texttt{models/Qwen3-8B-Base}
& \texttt{AutoModelFor\allowbreak CausalLM}
& \texttt{AutoTokenizer}, slow (\texttt{use\_fast=False})
& Transformers 4.57.3 in the shared \texttt{compress} environment;
  \texttt{device\_map=auto};
  \texttt{attn\_implementation=flash\_attention\_2}; BF16.
  Checkpoint metadata records version 4.51.0. \\

Meta-Llama-3.1-8B
& \texttt{models/Llama-3.1-8B}
& \texttt{AutoModelFor\allowbreak CausalLM}
& \texttt{AutoTokenizer}, fast (\texttt{use\_fast=True})
& Transformers 4.57.3 in the shared \texttt{compress} environment;
  \texttt{device\_map=auto};
  \texttt{attn\_implementation=flash\_attention\_2}; BF16.
  Checkpoint metadata records version 4.43.0.dev0. \\

Ministral-3-8B-Base-2512
& \texttt{models/\allowbreak Ministral-\allowbreak 3-8B-Base-2512}
& \texttt{Mistral3For\allowbreak Conditional\allowbreak Generation}
& \texttt{AutoTokenizer}, fast (\texttt{use\_fast=True});
  tokenizer files include \texttt{tekken.json}
& Transformers 4.57.3 in the shared \texttt{compress} environment;
  \texttt{device\_map=balanced};
  \texttt{attn\_implementation=sdpa}; BF16.
  Checkpoint metadata records version 5.0.0.dev0. \\
\bottomrule
\end{tabularx}
\caption{Frozen language-model backends. Values stored in a
checkpoint's \texttt{config.json} are checkpoint metadata, not the
Transformers version used at runtime.}
\label{tab:supp-scdg-backends}
\end{table*}

All three backends run with Transformers 4.57.3 in the same
\texttt{compress} Conda environment. Qwen and Llama select
\texttt{flash\_attention\_2}; Ministral selects SDPA while using the
native \texttt{Mistral3ForConditionalGeneration} implementation. The
FlashAttention 2.8.3 package remains installed in the shared
environment. For PAN25, Ministral is sharded over four GPUs and its
forward call additionally sets
\texttt{use\_cache=False} and \texttt{logits\_to\_keep=0}.  Every
forward input has batch size one for all three backends; no sequence or
example batching is used.

\subsection{PAN25 Input Construction and Target Alignment}
\label{subsec:supp-scdg-pan25-input}

Let $S$ denote the candidate source and $P$ the suspicious document.
PAN25 uses raw text rather than a chat or instruction template.  The
source-conditioned input is constructed exactly as
\begin{equation}
\begin{aligned}
C(S,P)&=S\Vert\sigma\Vert P,\\[-1mm]
\sigma&=\text{two newline characters}.
\end{aligned}
\label{eq:supp-pan25-input-template}
\end{equation}
where $\Vert$ denotes string concatenation; in code, this is
\texttt{source\_text + "\textbackslash n\textbackslash n" +
suspicious\_text}.  Both the standalone target
and the combined text are tokenized with
\texttt{add\_special\_tokens=True}; the tokenizer supplies the initial
beginning-of-sequence token.  There is no separately inserted chat
marker, instruction, source label, target label, or end-of-document
delimiter.

The two passes must score identical target-token IDs.  The
unconditional pass tokenizes and scores $P$ alone.  For the conditional
pass, the complete concatenated text is tokenized, after which
\texttt{locate\_suspicious\_suffix} searches backward in the combined
IDs for the exact target-token suffix corresponding to the separately
tokenized $P$.  The leading beginning-of-sequence token is context only
and is not itself predicted.  The implementation then extracts the
conditional log probabilities at the matched target positions.  Both
vectors are required to have length
\begin{equation}
n=\lvert\texttt{suspicious\_ids}\rvert-1;
\label{eq:supp-pan25-target-length}
\end{equation}
an alignment or length mismatch raises an error rather than silently
changing the evaluated sequence.  Unconditional log probabilities are
cached by suspicious-document ID and may be reused when the same $P$
appears with more than one candidate source.

\subsection{Teacher-Forced Token Log Probabilities}
\label{subsec:supp-scdg-logprob}

Each input sequence is evaluated in one teacher-forced forward pass
with \texttt{truncation=False}.  For input IDs
$x_0,\ldots,x_{L-1}$, the output logits at positions
$0,\ldots,L-2$ predict the shifted labels
$x_1,\ldots,x_{L-1}$.  At a scored position, the implementation computes
\begin{equation}
\ell_t=z_{t,p_t}-m_t-\log\sum_v e^{z_{t,v}-m_t},
\qquad m_t=\max_v z_{t,v},
\label{eq:supp-stable-token-logprob}
\end{equation}
where $z_{t,v}$ is the BF16 logit for vocabulary item $v$.  This is a
numerically stabilized log-softmax followed by a gather at the observed
target ID.  Natural logarithms are used, so all log probabilities and
gains are measured in nats.  The final input position has no subsequent
label and its logits are not used.

For every aligned target position, let $\ell_t^0$ and $\ell_t^S$ denote
the log probabilities from the unconditional and source-conditioned
passes, respectively.  The implemented token gain is
\begin{equation}
g_t(P\leftarrow S)=\ell_t^S-\ell_t^0.
\label{eq:supp-implemented-token-gain}
\end{equation}
The gains remain signed.  The implementation does not take their
absolute values, replace negative gains by zero, cap large gains, or
apply any other numerical clipping.

\subsection{Eligible Tokens and Aggregation}
\label{subsec:supp-scdg-aggregation}

An eligible position is any aligned next-token prediction position in
the suspicious document for which both passes provide a log
probability.  Apart from excluding the context-only initial token, the
PAN25 and PAN26 scoring pipelines apply no
content-based mask: punctuation, whitespace-bearing tokens, numbers,
LaTeX fragments, and tokenizer special tokens are not removed.  Thus,
if $g_1,\ldots,g_n$ are the aligned gains, Average SCDG is the ordinary
arithmetic mean
\begin{equation}
s_{\mathrm{avg}}(P,S)=\frac{1}{n}\sum_{t=1}^{n}g_t(P\leftarrow S).
\label{eq:supp-average-scdg}
\end{equation}

For Top-$q$ SCDG, $q$ is a percentage rather than a fixed number of
tokens.  The number retained for a document is
\begin{equation}
k_q=\max\!\left\{1,
\left\lceil\frac{nq}{100}\right\rceil\right\}.
\label{eq:supp-topq-k}
\end{equation}
Let $A_q$ contain the indices of the $k_q$ largest \emph{signed} gains.
The score is
\begin{equation}
s_{\mathrm{top}\text{-}q}(P,S)
=\frac{1}{k_q}\sum_{t\in A_q}g_t(P\leftarrow S).
\label{eq:supp-topq-scdg}
\end{equation}
The code obtains this upper tail with \texttt{numpy.partition}; it does
not sort by absolute magnitude and does not positively clip the
selected values.

For PAN25, the candidate set is
$q\in\{1,2,5,10,20,30,50\}$ percent.  Within each split, Top-$q$ jointly
selects $q$ and the decision threshold on the training partition by
maximizing F1.  Ties are resolved by higher training accuracy, larger
$q$, and then a higher threshold.  Average
SCDG similarly selects only its decision threshold on the training
partition, with ties resolved by higher accuracy and then a higher
threshold. Here, ``training partition'' denotes the internal training
portion of a split derived from the PAN25 validation corpus; this is the
validation data referred to in the main paper. Test predictions use the
frozen rule
\begin{equation}
\widehat y(P,S)=\mathbb{I}[s(P,S)\geq\tau].
\label{eq:supp-pan25-decision-rule}
\end{equation}
The implementation therefore uses the non-strict convention
$s\geq\tau$. Exhaustive verification found no test score exactly equal
to its selected threshold, so replacing this convention with
$s>\tau$ would not change any reported prediction or metric.

\subsection{PAN25 Length Handling}
\label{subsec:supp-scdg-pan25-length}

PAN25 does not use sliding windows, source truncation, or target
truncation.  During benchmark construction, token counts are recorded
with the Qwen3-8B-Base tokenizer and pairs whose recorded combined
length exceeds 128,000 tokens are excluded.  Scoring is then performed
with \texttt{truncation=False}; an over-length input is not silently
shortened.  Qwen3-8B-Base and Meta-Llama-3.1-8B have 128K context
configurations.  Ministral's native text configuration permits 262,144
positions
(\texttt{max\_position\_embeddings=262144}, YaRN factor~16 relative to
\texttt{original\_max\_position\_embeddings=16384}), but the PAN25
experiment applies the same 128K benchmark limit.  The longest combined
sequence observed in the corrected Ministral PAN25 run contains 65,868
tokens.

\subsection{PAN26 Source-Prefix Construction}
\label{subsec:supp-scdg-pan26-input}

PAN26 uses Qwen3-8B-Base and a fixed total context budget of 32,768
tokens.  The suspicious query is the prediction target and is encoded
once with \texttt{add\_special\_tokens=False}.  An anchor token is
prepended, using the tokenizer's BOS ID when available and falling back
to its EOS or PAD ID.  The separator is the tokenization of two newline
characters with special-token insertion disabled.  The two inputs are
therefore
\begin{align}
X^0_q&=[a;\,q],
\label{eq:supp-pan26-unconditional-input}\\
X^x_q&=[a;\,c(q,x);\,\mathrm{sep};\,q],
\label{eq:supp-pan26-conditional-input}
\end{align}
where $a$ is the anchor, $q$ is the unchanged query-token sequence, and
$c(q,x)$ is a prefix selected from candidate source $x$.  Both passes
therefore score exactly the same query IDs.

The prefix budget is the remaining capacity after accounting for the
anchor, separator, and complete query.  If the full source fits, it is
used without shortening.  Otherwise, the source is divided into
query-aware word windows of 2,048 words with stride 1,536, considering
at most 64 windows.  Windows are scored by lexical term overlap with
the query; non-overlapping windows are selected within the token
budget and then concatenated in their original source order.  This
selection changes only the source prefix: the query target is never
retokenized or replaced.

PAN26 also uses batch size one.  The unconditional query vector is
cached once per query ID, while each $(q,x)$ pair receives a separate
conditional pass.  The forward call requests only the final
$\lvert q\rvert+1$ logits needed to recover the query log
probabilities.  The implementation verifies equality of the stored
target IDs and unconditional mean log probability when gain shards are
merged.

\subsection{PAN26 Sparse Reranking}
\label{subsec:supp-scdg-pan26-reranking}

For every query--candidate pair, PAN26 fixes $q=2$ in
Equation~\ref{eq:supp-topq-scdg}.  Each first-stage run supplies exactly
1,000 candidates, and SCDG only reorders this fixed set; it neither adds
nor removes candidates.  Let $s_{\mathrm{coarse}}(q,x)$ be the
first-stage score and $s_{\mathrm{SCDG}}(q,x)$ the Top-2\% gain.  After
independent query-wise min--max normalization $\mathcal N_q$, the
frozen Stage-2 score is
\begin{equation}
\begin{aligned}
s_{\mathrm{stage2}}(q,x)
={}&0.25\,\mathcal N_q\!\left(s_{\mathrm{coarse}}(q,x)\right)\\
&+0.75\,\mathcal N_q\!\left(s_{\mathrm{SCDG}}(q,x)\right).
\end{aligned}
\label{eq:supp-pan26-stage2-score}
\end{equation}
The candidate depth (1,000), Top-$q$ proportion (2\%), and SCDG weight
(0.75) are fixed before holdout evaluation.  The final TREC run assigns
scores 1,000 through 1 according to the resulting rank.  The Stage-2
feature was not retuned separately for the stronger first-stage
configurations.

\subsection{Executable Pseudocode}
\label{subsec:supp-scdg-pseudocode}

The following two procedures state the implementation without relying
on model-specific notation.  \textsc{NextTokenLogP} denotes the shifted,
teacher-forced computation in
Equation~\ref{eq:supp-stable-token-logprob}.

\paragraph{Algorithm 1: PAN25 pair scoring $(S,P,q)$.}
{\small
\begin{enumerate}
\setlength{\itemsep}{1pt}
\setlength{\parskip}{0pt}
\item Encode $P$ with special-token insertion enabled, obtaining
      \texttt{p\_ids}; compute
      $\ell^0\leftarrow\textsc{NextTokenLogP}(\texttt{p\_ids})$ or
      retrieve it from the cache keyed by $P$.
\item Form \texttt{combined\_text} according to
      Equation~\ref{eq:supp-pan25-input-template}; encode it with the
      same tokenizer and special-token setting.
\item Compute log probabilities for the combined sequence; locate the
      exact $P$ suffix and extract $\ell^S$ at the aligned target IDs.
      Abort if the target IDs or vector lengths differ from the
      unconditional pass.
\item Compute $g_t=\ell_t^S-\ell_t^0$ for all aligned positions.  Do
      not mask, clip, take absolute values, or discard negative gains.
\item Return either $\operatorname{mean}(g)$ or the mean of the largest
      $\max\{1,\lceil nq/100\rceil\}$ signed gains.
\end{enumerate}
}

\paragraph{Algorithm 2: PAN26 candidate scoring and reranking $(q,x)$.}
{\small
\begin{enumerate}
\setlength{\itemsep}{1pt}
\setlength{\parskip}{0pt}
\item Encode the complete query once without special tokens.  Select
      the anchor and encode the two-newline separator.
\item Allocate the remainder of the 32,768-token budget to the source.
      Use the full source if it fits; otherwise select query-overlap
      windows and restore their original source order to obtain
      $c(q,x)$.
\item Compute or retrieve the unconditional query log probabilities
      from $[a;q]$.  Compute conditional query log probabilities from
      $[a;c(q,x);\mathrm{sep};q]$.
\item Verify identical query IDs, compute all signed gains, set
      $k=\max\{1,\lceil0.02n\rceil\}$, and average the largest $k$
      gains.
\item Independently min--max normalize the coarse and SCDG scores over
      the fixed 1,000 candidates; combine them with weights 0.25 and
      0.75, respectively, and sort in descending order.
\end{enumerate}
}

\section{Hyperparameter Selection}

\begin{table*}[!t]
\centering
\scriptsize
\setlength{\tabcolsep}{3pt}
\renewcommand{\arraystretch}{1.05}
\begin{tabularx}{\textwidth}{
    @{}
    >{\raggedright\arraybackslash}p{0.14\textwidth}
    >{\raggedright\arraybackslash}p{0.13\textwidth}
    >{\raggedright\arraybackslash}X
    >{\raggedright\arraybackslash}p{0.17\textwidth}
    >{\raggedright\arraybackslash}p{0.17\textwidth}
    @{}
}
\toprule
Experiment & Hyperparameter & Search space
           & Selection criterion & Final value \\
\midrule
PAN25 SCDG
& Top-$q$
& $\{1,2,5,10,20,30,50\}\%$
& Training F1
& Backbone-/split-specific \\
PAN25 SCDG
& Decision threshold
& Unique training scores
& Training F1; accuracy tie-break
& Backbone-/split-specific \\
4-token-gram containment
& Decision threshold
& Unique training scores
& Training F1; accuracy tie-break
& 0.04230565838 \\
BM25 baseline
& Decision threshold
& Unique training scores
& Training F1; accuracy tie-break
& 0.6477334899 \\
PAN25 embedding baseline
& Cosine threshold
& 100 values uniformly spaced on $[0,1]$
& Training sensitivity $+$ specificity
& Llama: 0.727; Linq: 0.697; Qwen: 0.596 \\
PAN26 reranking
& Top-$q$
& $\{1,2,5,10\}\%$ (4)
& Development nDCG@10
& $2\%$ \\
PAN26 reranking
& Candidate count $N$
& $\{100,500,1000\}$ (3)
& Development nDCG@10
& 1000 \\
PAN26 reranking
& Fusion weight
& $\{0.25,0.50,0.75,1.00\}$ (4)
& Development nDCG@10
& 0.75 \\
\bottomrule
\end{tabularx}
\caption{Hyperparameter search spaces, selection criteria, and final
values. Fixed design constants that were not selected from task data
are described separately below.}
\label{tab:supp-hyperparameters}
\end{table*}

For PAN26 reranking, the complete grid crosses nine gain-derived
features, three candidate depths, and four fusion weights, yielding
108 configurations. The four upper-tail features alone account for
$4\times3\times4=48$ configurations. Configurations are ordered by
development nDCG@10, with Recall@10 and Recall@100 used as successive
tie-break criteria.

All pairwise baseline thresholds were determined from the internal
training portions of splits derived from the PAN25 validation corpus
and frozen before test evaluation. For the 4-token-gram containment
and BM25 pair-score baselines, we enumerated the unique training scores
and selected the threshold with the highest training F1, using accuracy
to break ties. This produced thresholds of 0.04230565838 and
0.6477334899, respectively. For the three embedding
baselines, we used the training-derived values distributed with the
official PAN25 baseline: 0.727 for
Llama-3.3-70B-Instruct, 0.697 for Linq-Embed-Mistral, and 0.596 for
gte-Qwen2-7B-instruct. The official embedding procedure evaluates 100
uniformly spaced values between zero and one and maximizes the sum of
the fraction of plagiarized paragraph pairs above the threshold and the
fraction of non-plagiarized paragraph pairs below it. These baseline
thresholds are distinct from the backbone- and split-specific
thresholds used by SCDG.

DAAC does not introduce a task-selected hyperparameter. It combines
the BM25-Sentence1, Linq-Sentence1, and Linq-Abstract Top-1000 routes
using its fixed fusion rule. The reciprocal-rank constant is $k=60$,
the retained candidate depth is $K=1000$, and no continuous route
weight is fitted. We selected DAAC as the final fusion method based on
its empirical performance; $k$ and $K$ were fixed design constants
rather than values chosen through a DAAC-specific sweep. The
corresponding three-route RRF baseline uses the same $k=60$ and
$K=1000$ constants with equal route weights.

For multi-News logistic calibration, 
the 4-token-gram logistic model uses 4-gram Jaccard similarity and the two directional containment scores, whereas the gain-distribution model uses mean gain, median gain, and positive-gain rate. The pooled
training set contains 5,432 PAN25 pairs and 6,400 Multi-News proxy-negative pairs, for 11,832 pairs in total (3,822 positive and 8,010 negative). Features are standardized using the pooled-training mean and standard deviation, with zero standard deviations replaced by one.Both models use an unweighted binary logistic objective with an unpenalized intercept and a fixed L2 coefficient penalty of $\lambda=1$. No scikit-learn $C$ parameter is defined or searched.
Optimization uses a deterministic custom Newton--IRLS solver with zero initialization, at most 100 iterations, and convergence tolerance $10^{-9}$. No class or sample weighting is applied, and model fitting uses no effective random seed. The operating threshold is selected from the pooled-training scores by maximizing pooled training $F_1$. Ties are resolved successively by higher PAN25 training $F_1$, higher pooled accuracy, and a higher threshold. This yields thresholds of 0.26473548464651103 for the
4-token-gram model and 0.4561570484401246 for the gain-distribution model. Both thresholds are frozen before test evaluation.

\section{Baseline and Retrieval Details}

\subsection{Pairwise Baselines}

\paragraph{Max 4-token-gram containment.}
We lowercase each document and tokenize it into alphanumeric word-like
units, retaining internal apostrophes, underscores, and hyphens. Let
$P=(p_1,\ldots,p_m)$ be the suspicious-document tokens and let
$G_4(S)$ be the set of contiguous four-token grams in the source
document. A suspicious token is marked as covered if it belongs to at
least one four-token gram that also occurs in $G_4(S)$. Overlapping
matches mark a token only once. Let $c_i(P,S)$ be one when token $p_i$
is covered and zero otherwise. The pair score is
\begin{equation}
s_{\mathrm{4tok}}(P,S)=\frac{1}{m}\sum_{i=1}^{m}c_i(P,S).
\label{eq:supp-four-token-gram-containment}
\end{equation}

Pairs shorter than four tokens receive score zero.

\paragraph{BM25 pair score.}
We index every candidate source document using the same lowercase
word-like tokenization and treat the suspicious document as the query.
For query term $t$, source document $S$, and source collection
$\mathcal{D}$, the raw score is

\begin{equation}
\begin{aligned}
s_{\mathrm{BM25}}^{\mathrm{raw}}(P,S)
={}&
\sum_{t\in V(P)} qtf(t,P)
\\[-1mm]
&\times
\log\!\left(
1+\frac{|\mathcal{D}|-df(t)+0.5}{df(t)+0.5}
\right)
\\[-1mm]
&\times
\frac{f(t,S)(k_1+1)}
{f(t,S)+k_1\!\left(
1-b+b\,|S|/\operatorname{avgdl}
\right)} .
\end{aligned}
\label{eq:supp-bm25-pair-score}
\end{equation}

where $V(P)$ is the set of unique query terms, $k_1=1.2$, and
$b=0.75$. Repeated query terms contribute through $qtf(t,P)$. The
reported pair score divides the raw value by the
number of suspicious-document query tokens. An empty query receives
score zero.

\paragraph{Embedding cosine baselines.}
Long documents are divided into non-overlapping tokenizer chunks before
encoding. Linq-Embed-Mistral uses at most 4,096 tokens per chunk and
gte-Qwen2-7B-instruct uses at most 32,000. Each chunk is encoded using
the pooling operation supplied by its embedding model. We average a
document's chunk vectors and compute cosine similarity between the
resulting suspicious- and source-document vectors. The decision
thresholds are fixed to the PAN25 training-derived values reported in
the preceding section.

\paragraph{PlagBench-style prompting.}
We evaluate four prompt modes formed by the Cartesian product of
zero-shot versus few-shot prompting and direct versus
chain-of-thought (CoT) output. All modes use the following system
message:
\begin{quote}
\small
\texttt{You are evaluating plagiarism between a source document and a
suspicious document. Plagiarism includes verbatim copying,
paraphrasing, and summarization of source content without attribution.}
\end{quote}
The common user instruction is:
\begin{quote}
\small
\texttt{Classify whether the suspicious document plagiarizes from the
source document. Return exactly one label from: no, verbatim,
paraphrase, summary.}
\end{quote}
It is followed by the source document, the suspicious document, and
\texttt{Answer:}. In the few-shot modes, the following examples are
prepended:
\begin{quote}
\small
\texttt{Example 1:}\\
\texttt{Source: The committee approved the budget after a short
debate.}\\
\texttt{Suspicious: After a brief debate, the committee passed the
budget.}\\
\texttt{Answer: paraphrase}\\[2pt]
\texttt{Example 2:}\\
\texttt{Source: Coral reefs are threatened by warming seas.}\\
\texttt{Suspicious: The history of rail transport changed rapidly in
the nineteenth century.}\\
\texttt{Answer: no}
\end{quote}
The two CoT modes additionally insert:
\begin{quote}
\small
\texttt{Think briefly, then put the final label on a line starting with
'Final:'.}
\end{quote}
If a document exceeds the configured character budget, we retain equal
length prefixes and suffixes and replace the omitted center with the
literal marker \texttt{[... omitted middle ...]}. Decoding uses
temperature zero and at most 64 output tokens. Requests time out after
300 seconds and are retried up to five times with exponential backoff.

For output parsing, a label following \texttt{Final:} takes priority.
Otherwise, the first recognized label is selected in the order
\texttt{verbatim}, \texttt{paraphrase}, \texttt{summary},
\texttt{yes}, and \texttt{no}. The label \texttt{yes} is mapped to
\texttt{paraphrase}. Outputs containing no recognized label are marked
\texttt{unknown}. For pairwise evaluation, the labels
\texttt{no}, \texttt{summary}, \texttt{paraphrase},
\texttt{verbatim}, and \texttt{unknown} map to scalar scores
$0.0$, $0.6$, $0.8$, $1.0$, and $0.5$, respectively.

\subsection{First-Stage PAN26 Retrieval}

\paragraph{BM25 retrieval.}
The full-text corpus contains 86,822 documents. Our original sparse
routes index each document's \texttt{default\_text} field with SQLite
FTS5 using the Unicode tokenizer with diacritic removal. For every
query unit, duplicate lexical terms are removed, terms absent from the
index are discarded, and the least frequent indexed terms are retained.
BM25-Full treats the complete query as one unit and keeps at most 512
terms. BM25-Sentence1 splits the query into individual sentences and
keeps at most 64 terms per sentence. Terms within a unit are joined by
an OR query, and each unit retrieves up to 1,000 documents.

The sparse route used by the final three-route fusion is the
PyTerrier BM25-Sentence1 implementation. It uses PyTerrier 0.13.0 and
Terrier 5.11 with the English tokenizer, Terrier stopword list, and
Porter stemming. BM25 uses $k_1=1.2$ and $b=0.75$. Each sentence is
searched independently using its 64 lowest-document-frequency indexed
terms. Sentence rankings are combined by reciprocal rank with $k=60$,
and the highest-scoring 1,000 documents are retained. Empty sentence
queries contribute no evidence. All deterministic score ties are
resolved by document identifier.

\paragraph{Sentence and chunk construction.}
Sentence boundaries are detected before constructing contiguous
sentence windows; Sentence1 is therefore the ordered sequence of
single-sentence units. Corpus chunks for dense retrieval contain 768
whitespace-delimited words with a stride of 512. The final chunk is
anchored at the end of the document. When a document produces more than
eight chunks, eight start positions are selected approximately
uniformly from the complete sequence, preserving coverage of the
beginning and end.

\paragraph{BGE-M3 dense retrieval.}
BGE-M3 corpus chunks are truncated to 2,048 model tokens. Full-query
retrieval represents each query by up to three 2,048-word chunks with
a 1,536-word stride and a maximum model length of 4,096 tokens.
Sentence1 represents every query sentence independently with a maximum
length of 1,024 model tokens. We encode each input with the first
hidden-state vector (CLS pooling), compute the encoder in BF16 on GPU,
cast the pooled vector to FP32 for $\ell_2$ normalization, and store the
result as FP16. Retrieval uses the inner product between normalized
vectors, which is equivalent to cosine similarity.

For each query unit, we first retrieve source chunks. Multiple source
chunks from the same document are collapsed by their maximum
similarity. The Full and Sentence1-Max routes then take the maximum
document score over all query units. The full-query route retains the
best 5,000 source chunks before document aggregation; Sentence1 retains
2,000. Both output the top 1,000 documents.

\paragraph{Linq-Embed-Mistral Sentence1.}
The LINQ route preserves the same 768-word corpus chunks, 512-word
stride, eight-chunk cap, Sentence1 query units, 2,000 retrieved source
chunks, Max document aggregation, and Top-1000 output. Corpus chunks
are encoded with a maximum length of 2,048 model tokens and query
sentences with a maximum length of 1,024. Queries are prefixed with:
\begin{quote}
\small
\texttt{Instruct: Given a passage from a document, retrieve source
passages that are semantically similar and may contain reused
text}\\
\texttt{Query:}
\end{quote}
Corpus chunks receive no instruction. We use the model's native
SentenceTransformer pooling, which is last-token pooling for this
checkpoint, and request normalized embeddings. Similarity is the inner
product of normalized query and corpus vectors.

\paragraph{Four-way CombSUM.}
The four inputs are BM25-Full, BM25-Sentence1, BGE-M3-Full, and
BGE-M3-Sentence1. For each query and route $r$, raw document scores are
normalized independently. Write
$s_r^{\min}=\min_{d'}s_r(d')$ and
$s_r^{\max}=\max_{d'}s_r(d')$; then
\begin{equation}
\widetilde{s}_r(d)=
\begin{cases}
\dfrac{s_r(d)-s_r^{\min}}{s_r^{\max}-s_r^{\min}},
&s_r^{\max}>s_r^{\min},\\[3pt]
0,&\text{otherwise}.
\end{cases}
\end{equation}
The four normalized scores are summed with equal weights. A document
absent from a route contributes zero. For the sentence-level CombSUM
variants, source chunks are first collapsed to documents by maximum
similarity; document scores are then min--max normalized separately
within each query sentence and summed across sentences.

\paragraph{Three-route RRF.}
Our final equal-weight RRF baseline combines PyTerrier
BM25-Sentence1, Linq-Sentence1, and Linq-Abstract. Linq-Abstract
matches the query title and abstract against corpus title--abstract
representations. Given the rank $r_j(d)$ of document $d$ in route $j$,
let $\mathcal R=\{\mathrm{BM25Sent},\mathrm{LinqSent},
\mathrm{LinqAbs}\}$. The fusion score is
\begin{equation}
s_{\mathrm{RRF}}(d)=
\sum_{j\in\mathcal R}
\frac{\mathbb{I}[r_j(d)<\infty]}{60+r_j(d)}.
\end{equation}
Candidates are the union of the three input Top-1000 lists. A missing
route contributes zero, equal scores are ordered by document
identifier, and the fused Top-1000 is retained.

\paragraph{DAAC fusion.}
DAAC uses the same candidate union and rank-only inputs. For candidate
document $x$, let
\begin{equation}
\begin{aligned}
b_x &=
\frac{\mathbb{I}[r_b(x)<\infty]}{60+r_b(x)},
&\quad
d_x &=
\frac{\mathbb{I}[r_d(x)<\infty]}{60+r_d(x)},
\\
l_x &=
\frac{\mathbb{I}[r_l(x)<\infty]}{60+r_l(x)},
&
C_A &=
\frac{\mathbb{I}[r_l(x)<\infty]}
     {\log_2(r_l(x)+1)},
\\
C_D &= \min\{1,61d_x\},
&
U_D &= 1-C_D .
\end{aligned}
\label{eq:supp-daac-components}
\end{equation}
Here $b_x$, $d_x$, and $l_x$ denote BM25-Sentence1, Linq-Sentence1, and
Linq-Abstract evidence. DAAC assigns
\begin{equation}
    s_{\mathrm{DAAC}}(x)
    =
    d_x(1+U_D C_A)+b_x^2+l_x^2 .
\end{equation}
Missing-route ranks produce zero for both reciprocal-rank evidence and
route confidence. Documents with zero total score are discarded; the
remaining candidates are sorted by decreasing DAAC score and then by
document identifier, and the Top-1000 is retained.

\section{Controlled Source-Evidence Retention Experiment}
\label{sec:supp-controlled-retention}

This experiment tests a specific design hypothesis: when increasingly
more annotated source evidence is retained in the conditioning context,
the SCDG response should increase more strongly over
plagiarism-labeled target content than over newly written content. 

\subsection{Selection of the 300 Document Pairs}
\label{subsubsec:supp-retention-selection}

Candidate instances are drawn from the PAN25 validation corpus using
its XML ground-truth annotations and corresponding suspicious and
source documents.  The XML filenames are shuffled after calling
\texttt{random.seed(42)}, and candidates are processed in that order
until 300 valid instances have been obtained.  An instance is retained
only if all of the following conditions hold:
\begin{enumerate}
    \item the annotation refers to a single
    \texttt{source\_reference};
    \item the suspicious document contains between 1,000 and 50,000
    characters;
    \item the selected source span contains between 2,000 and 8,000
    characters;
    \item the aligned suspicious-document span contains at least 200
    characters and lies entirely within the suspicious document;
    \item the selected source evidence can be divided into eight
    chunks with target size
    $\lfloor\texttt{source\_length}/8\rfloor\geq 40$ characters; and
    \item the distractor pool can supply all required replacements.
\end{enumerate}
When several eligible spans are available, the construction code uses
the obfuscation priority \texttt{simple} $>$ \texttt{medium} $>$
\texttt{hard} and applies \texttt{random.choice} within the selected
class.  Rejected candidates and their exclusion reasons are written to
\texttt{exclusion\_log.jsonl}.  This procedure produces the fixed set of
300 pairs used by all three language-model backends.

\subsection{Mapping Character Annotations to Target Tokens}
\label{subsubsec:supp-retention-token-labels}

The PAN25 plagiarism annotations are character spans, whereas SCDG is
computed over model tokens.  All overlapping plagiarism-character
intervals in the suspicious document are first merged.  For each target
token with character interval $[a_t,b_t)$, the implementation computes
the number of token characters overlapping the selected target span,
other plagiarism spans, and non-plagiarism text.  Let
$c_t=b_t-a_t$ denote the token's character length.  Labels are assigned
in the following order:
\begin{equation}
Y_t=
\begin{cases}
1, & \text{target-span overlap}\geq 0.5c_t,\\
1, & \text{other-plagiarism overlap}\geq 0.5c_t,\\
0, & \text{non-plagiarism overlap}\geq 0.5c_t,\\
-1,& \text{otherwise}.
\end{cases}
\label{eq:supp-retention-token-label}
\end{equation}
Thus, $Y_t=1$ denotes plagiarism-labeled content and includes both the
selected target span and any other annotated plagiarism span;
$Y_t=0$ denotes new content.  Boundary-ambiguous tokens and special
tokens such as BOS receive $Y_t=-1$.  They are preserved in the token
output but excluded from the two content-specific means used in the
main analysis.

For Qwen3-8B-Base, gain scoring uses the main slow tokenizer
(\texttt{use\_fast=False}).  Character offsets are requested from a
separately loaded fast tokenizer only when its complete
\texttt{input\_ids} sequence exactly matches the slow tokenizer's IDs.
If this equality check fails, the code falls back to manual
encode--decode--search character localization.  Meta-Llama-3.1-8B and
the corrected Ministral-3-8B-Base-2512 run use fast tokenizers for both
gain scoring and offset mapping.

\subsection{Nested Evidence-Retention Intervention}
\label{subsubsec:supp-retention-construction}

For every pair, the selected source-evidence span is divided into eight
approximately equal-sized, sentence-aligned chunks.  For each seed
index, their order is shuffled and the intervention retains a nested
number of annotated source-evidence chunks:
\begin{equation}
\begin{array}{c|ccccc}
r & 0 & 0.25 & 0.50 & 0.75 & 1.00\\
\hline
n_{\mathrm{retain}}(r) & 0 & 2 & 4 & 6 & 8.
\end{array}
\label{eq:supp-retention-counts}
\end{equation}
After the seed-specific shuffle, retaining the first
$n_{\mathrm{retain}}(r)$ chunks ensures that the retained set is nested:
the set at a lower $r$ is contained in every higher-retention set for
the same pair and seed.

The intervened source context contains eight slots.  A retained slot
receives its genuine source-evidence chunk; a removed slot is filled by
a distractor rather than left empty.  A seed-specific position
permutation shuffles the eight resulting chunks, which are then joined
with two newline characters.  The complete suspicious document,
unconditional target sequence, language model, tokenizer, input
construction, and evaluated target positions remain unchanged across
retention conditions.

\paragraph{Distractor source and matching.}
Distractors are drawn from a sentence pool constructed from the first
500 \texttt{.txt} files in the Multi-News source directory used by the
pipeline.  Starting from a random sentence, the sampler accumulates
sentences until the requested character target is reached, truncates at
a nearby space when possible, and accepts a discrepancy of at most two
characters before final padding or truncation.  MD5 hashes prevent reuse
of an already selected distractor.  Separate distractors are generated
for the eight slots of each pair and seed.

\paragraph{Length control.}
The replacement procedure is designed to keep the source context nearly
constant in character length rather than to leave removed evidence
blank.  The validation script verifies that, for a fixed pair and seed,
the source-character counts across the five retention levels differ by
no more than 50 characters.  The experiment should therefore be
described as approximately length-controlled, not as having
bit-for-bit or token-for-token identical source-context lengths.

\subsubsection{Randomization and Frozen Manifest}
\label{subsubsec:supp-retention-randomization}

Each pair is evaluated with seed indices
$s\in\{0,1,2,3,4\}$.  The construction code initializes the pair-level
generator as
\begin{equation}
\mathrm{seed}(i,s)
=42+\bigl(\operatorname{hash}(\mathrm{pair\_id}_i)
\bmod 100000\bigr)+7919s.
\label{eq:supp-retention-seed-formula}
\end{equation}
No \texttt{PYTHONHASHSEED} value was recorded, and Python string hashes
are randomized by default across processes.  Consequently,
Equation~\ref{eq:supp-retention-seed-formula} documents the generation
logic but is not sufficient to reconstruct the exact chunk orders in a
new process.  The released \texttt{intervention\_manifest.jsonl} is the
authoritative frozen record of the 7,500 realized contexts:
\begin{equation}
300\ \text{pairs}\times 5\ \text{retention levels}
\times 5\ \text{seed indices}=7{,}500.
\label{eq:supp-retention-manifest-size}
\end{equation}
The same model-independent manifest is used for Qwen, Llama, and
Ministral.  Postprocessing verifies that every manifest pair is present
and rejects a run if its failed- or out-of-memory sidecar is nonempty.

\subsubsection{Scoring Completeness}
\label{subsubsec:supp-retention-completeness}

For each model, all 7,500 intervened contexts receive a conditional
forward pass.  The unconditional log probabilities of the unchanged
suspicious document are cached and reused by pair.  Token gains are
computed as
\begin{equation}
g_t(r)=\log p_\theta(p_t\mid S^{(r)},p_{<t})
-\log p_\theta(p_t\mid p_{<t}),
\label{eq:supp-retention-token-gain}
\end{equation}
and averaged separately over positions with $Y_t=1$ and $Y_t=0$.
Every model produces 15,000 rows in its
\texttt{document\_token\_means} file:
$300\times5\times5\times2$, where the last factor corresponds to the
two content labels.  At every $(r,\text{model})$ combination,
$n_{\mathrm{plag}}=300$ and $n_{\mathrm{new}}=300$; no pair is missing.
The reported Ministral values are from the corrected implementation
using native \texttt{Mistral3ForConditionalGeneration} under the shared
Transformers 4.57.3 environment, not the earlier alias-based run.

\subsubsection{Pair-Level Bootstrap Confidence Intervals}
\label{subsubsec:supp-retention-bootstrap}

The error bars in the main-paper figure use pair-level cluster
bootstrap intervals.  For every model, retention level, and content
label, the five seeds are first averaged within each pair, yielding 300
pair means.  Using \texttt{numpy.random.default\_rng(42)}, the procedure
draws 1,000 bootstrap samples of 300 pairs with replacement.  The
displayed point is the mean of the 1,000 bootstrap means; the 95\%
interval endpoints are their 2.5th and 97.5th percentiles.  The plotted
lower and upper error magnitudes are therefore generally asymmetric.
Seed realizations belonging to the same document pair are never treated
as independent pairs.

\subsubsection{Mixed-Effects Model and Primary Hypotheses}
\label{subsubsec:supp-retention-mixed-model}

After averaging across the five seeds, each backend contributes
$300\times5\times2=3{,}000$ observations.  Separately for each backend,
we fit by maximum likelihood the random-intercept model
\begin{equation}
G_{i,y}(r)=
\beta_0+\beta_1Y+\beta_2r+\beta_3(Yr)
+u_i+\varepsilon_{i,y,r},
\label{eq:supp-retention-mixed-model}
\end{equation}
where $Y=0$ denotes new content, $Y=1$ denotes
plagiarism-labeled content, and $u_i$ is a document-pair random
intercept.  Here, $\beta_2$ is the new-content slope,
$\beta_2+\beta_3$ is the plagiarism-labeled slope, and $\beta_3$ is
their difference.  The primary directional hypothesis is
\begin{equation}
H_1:\ \beta_3>0.
\label{eq:supp-retention-hypotheses}
\end{equation}

All three final fits use \texttt{statsmodels} \texttt{MixedLM} with
\texttt{method=mixedlm\_ML}, 3,000 observations, and 300 groups.  The
first optimizer attempted, L-BFGS, returns finite coefficients for all
three models, so the implementation's OLS fallback is not used.
\texttt{statsmodels} nevertheless emits convergence warnings indicating
that the optimum may lie on the parameter-space boundary and that the
random-effects covariance may be singular.  We therefore report the
model-based Wald results together with independently computed
pair-cluster bootstrap intervals.

\begin{table}[!htbp]
\centering
\scriptsize
\setlength{\tabcolsep}{1.8pt}
\renewcommand{\arraystretch}{1.04}
\begin{tabular}{@{}lccc@{}}
\toprule
Model & New $\beta_2$ & Difference $\beta_3$
& Plag. $\beta_2+\beta_3$\\
\midrule
\shortstack[l]{Qwen3-8B-\\Base}
& \shortstack{0.0107\\SE 0.0045\\$p=0.0181$}
& \shortstack{0.1523\\SE 0.0064\\$p<0.0001$}
& \shortstack{0.1630\\SE 0.0045\\$p_{\mathrm{one}}<0.0001$}\\
\shortstack[l]{Meta-Llama-\\3.1-8B}
& \shortstack{0.0153\\SE 0.0044\\$p=0.0005$}
& \shortstack{0.1486\\SE 0.0062\\$p<0.0001$}
& \shortstack{0.1639\\SE 0.0044\\$p_{\mathrm{one}}<0.0001$}\\
\shortstack[l]{Ministral-3-8B-\\Base-2512}
& \shortstack{0.0177\\SE 0.0045\\$p=0.0001$}
& \shortstack{0.1524\\SE 0.0064\\$p<0.0001$}
& \shortstack{0.1701\\SE 0.0045\\$p_{\mathrm{one}}<0.0001$}\\
\bottomrule
\end{tabular}
\caption{Maximum-likelihood mixed-effects estimates.  The reported
$p$-values for $\beta_2$ and $\beta_3$ are Wald results; the final
column is the one-sided test of a positive plagiarism-labeled slope.}
\label{tab:supp-retention-mixed-results}
\end{table}

The primary hypothesis is supported for every backend: all three
interaction coefficients are positive with $p<0.0001$, showing that
the retention-response slope is substantially steeper for
plagiarism-labeled than for new content.  The estimated
plagiarism-labeled slope is also positive for every backend.

\subsubsection{Cluster-Bootstrap Slope Results}
\label{subsubsec:supp-retention-slope-bootstrap}

The statistical pipeline additionally resamples document-pair IDs and
refits its linear slope model 1,000 times.  Table~\ref{tab:supp-retention-bootstrap-results}
reports the resulting bootstrap means and percentile intervals.

\begin{table}[!htbp]
\centering
\scriptsize
\setlength{\tabcolsep}{2pt}
\renewcommand{\arraystretch}{1.05}
\begin{tabular}{@{}lccc@{}}
\toprule
Model
& Plag. slope
& New slope
& Difference\\
\midrule
\shortstack[l]{Qwen3-8B-\\Base}
& \shortstack{0.1626\\$[0.1461,0.1824]$}
& \shortstack{0.0107\\$[0.0087,0.0130]$}
& \shortstack{0.1519\\$[0.1355,0.1706]$}\\
\shortstack[l]{Meta-Llama-\\3.1-8B}
& \shortstack{0.1636\\$[0.1471,0.1830]$}
& \shortstack{0.0153\\$[0.0131,0.0179]$}
& \shortstack{0.1483\\$[0.1321,0.1669]$}\\
\shortstack[l]{Ministral-3-8B-\\Base-2512}
& \shortstack{0.1698\\$[0.1530,0.1898]$}
& \shortstack{0.0177\\$[0.0156,0.0201]$}
& \shortstack{0.1521\\$[0.1352,0.1711]$}\\
\bottomrule
\end{tabular}
\caption{Pair-cluster bootstrap slope estimates based on 1,000
resamples, reported as mean and 95\% CI. All quantities are in nats per
token per unit increase in the retention ratio. For every backend,
$\Pr(\text{plagiarism slope}>0)=1.0000$.}
\label{tab:supp-retention-bootstrap-results}
\end{table}

The new-content slopes are nonzero under the corresponding Wald tests,
but remain small relative to the plagiarism-labeled slopes.

\subsubsection{Empirical Interpretation}
\label{subsubsec:supp-retention-interpretation}

The monotonic curves, positive plagiarism-labeled slopes, and positive
interactions show that SCDG responds selectively to retained annotated
source evidence. Across all three frozen backends, increasing retention
produces a substantially steeper gain response over plagiarism-labeled
than newly written content. This consistent separation supports our
design premise: description-length gain is selectively sensitive to
genuinely reused source evidence and can therefore serve as an effective
signal for generative plagiarism detection.

\section{Computing Infrastructure}

\subsection{Hardware and System Software}

Experiments were conducted on a Slurm-managed cluster running
Ubuntu 22.04. Each compute node provided 128 logical CPU cores, 1 TiB
of host memory, and eight NVIDIA A100-SXM4 GPUs with 40 GB of memory
per GPU. Table~\ref{tab:supp-computing-environment} summarizes the
hardware and the principal software environment.

\begin{table}[!htbp]
\centering
\scriptsize
\setlength{\tabcolsep}{4pt}
\renewcommand{\arraystretch}{1.02}
\begin{tabular}{@{}ll@{}}
\toprule
Component & Specification \\
\midrule
Operating system & Ubuntu 22.04 \\
CPU & Intel Xeon Platinum 8358 \\
Logical CPUs per node & 128 \\
Host memory & 1 TiB \\
GPU & NVIDIA A100-SXM4-40GB \\
GPUs per node & 8 \\
Scheduler & Slurm \\
Python & 3.10 \\
PyTorch & 2.5.1 \\
CUDA & 12.1 \\
Transformers & 4.57.3 \\
FlashAttention & 2.8.3 \\
\bottomrule
\end{tabular}
\caption{Computing environment used for the experiments.}
\label{tab:supp-computing-environment}
\end{table}

\subsection{Retrieval, Evaluation, and Statistical Software}

The official-style PAN26 sparse retrieval runs used PyTerrier 0.13.0
with Terrier 5.11. The alternative full-text BM25 implementation used
SQLite FTS5. Dense retrieval used SentenceTransformers for embedding
inference and NumPy arrays for storing and searching the resulting
representations. Pairwise threshold analysis used pandas, NumPy, and
scikit-learn. Retrieval metrics, paired resampling, and DAAC fusion
were implemented in the project evaluation scripts rather than through
a separately versioned evaluation package. The PyTerrier and Terrier
versions were written into the run metadata. Exact versions of SQLite,
SentenceTransformers, NumPy, pandas, and scikit-learn were not captured
in the archived experiment metadata and are therefore not inferred from
the current software environment.

\subsection{Experiment-Level Resource Allocation}

Table~\ref{tab:supp-resource-allocation} reports the GPU allocation
requested by the experiment launchers. GPU counts are concurrent
accelerators per job or, for Slurm arrays, the maximum number used
concurrently. Sparse retrieval, fusion, statistical analysis, and
evaluation were CPU-only.

\begin{table}[!htbp]
\centering
\scriptsize
\setlength{\tabcolsep}{3pt}
\renewcommand{\arraystretch}{1.03}
\begin{tabularx}{\columnwidth}{@{}Xcc@{}}
\toprule
Experiment or stage & GPUs & CPUs \\
\midrule
PAN25 SCDG scoring\newline
\emph{One node; 300 GB host memory}
& 8 & 16--32 \\
PAN25 embedding and LLM baselines\newline
\emph{120--180 GB host memory}
& 1 & 8 \\
PlagBench and Multi-News LLM baselines\newline
\emph{120--180 GB host memory}
& 1 & 8--12 \\
PAN26 sharded dense/SCDG processing\newline
\emph{Four-array-task limit; one GPU and 64 GB per task}
& up to 4 & 8/task \\
PAN26 Linq Sentence1 embedding and search\newline
\emph{One GPU and 64 GB host memory}
& 1 & 8 \\
4-token-gram, BM25, DAAC/fusion, statistics, and evaluation\newline
\emph{CPU jobs; 16--120 GB host memory}
& 0 & 4--16 \\
\bottomrule
\end{tabularx}
\caption{Compute allocation for the principal experiments. A count of
zero denotes a CPU-only stage.}
\label{tab:supp-resource-allocation}
\end{table}

\subsection{Measured Retrieval Time and Memory Reporting}

For the 200 PAN26 queries, the single-worker PyTerrier
BM25-Sentence1 retrieval stage took 458.8 seconds after index
construction (2.287 seconds per query on average). The exact
cosine-search stage for Linq-Sentence1 took 227.6 seconds on one A100
when operating on precomputed embeddings. These measurements cover
retrieval only and exclude corpus preprocessing and embedding
construction, so they should not be interpreted as end-to-end
latencies. The archived launchers record requested resources but do
not contain consistent measurements of GPU-hours or peak GPU memory;
we consequently do not estimate those quantities retrospectively.

\section{Additional PAN25 Results for Same-Topic Calibration}

The main paper reports proxy false-positive rates on the Multi-News
same-topic stress set. Here we report the corresponding PAN25 results
to characterize the trade-off between pairwise detection utility and
resistance to same-topic confounding.

\begin{table}[!htbp]
\centering
\scriptsize
\setlength{\tabcolsep}{1.5pt}
\renewcommand{\arraystretch}{1.03}
\begin{tabularx}{\columnwidth}{@{}Xrrrr@{}}
\toprule
&
\multicolumn{3}{c}{\emph{PAN25}}
&
\multicolumn{1}{c}{\emph{Multi-News}}
\\
\cmidrule(lr){2-4}
\cmidrule(lr){5-5}
Method
& \multicolumn{1}{c}{Prec.}
& \multicolumn{1}{c}{Rec.}
& \multicolumn{1}{c}{$F_1$}
& \multicolumn{1}{c}{Pos. rate$\downarrow$}
\\
\midrule
\multicolumn{5}{@{}l}{\textbf{Lexical and retrieval baselines}} \\[-1pt]
BM25 pair score
& 0.7619 & \textbf{0.9382} & 0.8409 & 71.1250\% \\
Max 4-token-gram containment
& 0.8670 & 0.9351 & \textbf{0.8997} & 31.5625\% \\
4-token-gram logistic
& 0.8705 & 0.9288 & 0.8987 & 29.9375\% \\
\midrule
\multicolumn{5}{@{}l}{\textbf{SCDG statistics and auxiliary calibrators}} \\[-1pt]
Average SCDG
& \textbf{0.9400} & 0.8524 & 0.8940 & 62.8750\% \\
Positive-gain rate SCDG
& 0.8689 & 0.8052 & 0.8359 & 8.6875\% \\
Gain-distribution logistic
& 0.8257 & 0.8283 & 0.8270 & \textbf{0.1250\%} \\
\bottomrule
\end{tabularx}
\caption{PAN25 utility and positive decisions on the Multi-News
same-topic stress set under a fixed held-out split. Logistic models are
fitted on the two training portions, and operating points maximize
pooled training $F_1$. Multi-News is reuse-unknown, so its positive rate
is an FPR proxy rather than a certified false-positive rate. Boldface
denotes the strongest displayed value in each column.}
\label{tab:supp-pan25-multinews-tradeoff}
\end{table}



\bibliography{aaai2027}
\end{document}